\documentclass[10pt,journal,compsoc]{IEEEtran}
\usepackage{amsmath,amsfonts}
\usepackage{array}
\usepackage[caption=false,font=normalsize,labelfont=sf,textfont=sf]{subfig}
\usepackage{textcomp}
\usepackage{stfloats}
\usepackage{url}
\usepackage{verbatim}
\usepackage{graphicx}
\usepackage{times}
\usepackage{soul}
\usepackage{url}
\usepackage[hidelinks]{hyperref}
\usepackage{graphicx}
\usepackage{amsmath}
\usepackage{amsthm}
\usepackage{booktabs}
\usepackage{subfig}
\usepackage{bbold}
\usepackage{cases}
\usepackage{multirow}
\usepackage{bbding}
\usepackage{pifont}
\usepackage{subfig,graphicx}
\usepackage{makecell}
\usepackage{algorithm, algorithmicx, algpseudocode}

\ifCLASSOPTIONcompsoc
  \usepackage[nocompress]{cite}
\else
  \usepackage{cite}
\fi
\ifCLASSINFOpdf
\else
\fi
\begin{document}

%
\title{Beyond Anti-Forgetting: Multimodal Continual Instruction Tuning with Positive Forward Transfer}
%
%
%
%

\author{Junhao~Zheng, Qianli~Ma$^{*}$,~\IEEEmembership{Member,~IEEE,} Zhen~Liu, Binquan~Wu, Huawen~Feng
\thanks{$^{*}$Corresponding author: Qianli Ma.}
\thanks{The authors are with the School of Computer Science and Engineering, South China University of Technology, Guangzhou 510006, China (e-mail: junhaozheng47@outlook.com; qianlima@scut.edu.cn; cszhenliu@mail.scut.edu.cn; cskyun\_ng@mail.scut.edu.cn; 541119578@qq.com).}
\thanks{Manuscript received April 19, 2005; revised August 26, 2015.}}
\markboth{Journal of \LaTeX\ Class Files,~Vol.~14, No.~8, August~2015}%
{Zheng \MakeLowercase{\textit{et al.}}: Beyond Anti-Forgetting: Multimodal Continual Instruction Tuning with Positive Forward Transfer}
%



\IEEEtitleabstractindextext{%
\begin{abstract}
Multimodal Continual Instruction Tuning (MCIT) enables Multimodal Large Language Models (MLLMs) to meet continuously emerging requirements without expensive retraining. MCIT faces two major obstacles: catastrophic forgetting (where old knowledge is forgotten) and negative forward transfer (where the performance of future tasks is degraded). Although existing methods have greatly alleviated catastrophic forgetting, they still suffer from negative forward transfer. We discover a large discrepancy in different input embeddings by performing singular value decomposition (SVD) on input embeddings. This discrepancy results in the model learning irrelevant information for old and pre-trained tasks, leading to catastrophic forgetting and negative forward transfer. To address these issues, we propose Prompt Tuning with Positive Forward Transfer (Fwd-Prompt), a prompt-based method that projects the prompt gradient to the residual space to minimize interference between tasks and to the pre-trained subspace for reusing pre-trained knowledge. Our experiments demonstrate that Fwd-Prompt achieves state-of-the-art performance while updating fewer parameters and requiring no old samples. Our research illuminates the potential of continuously adapting MLLMs to new tasks under the instruction tuning paradigm and encourages future studies to explore MCIT.
\end{abstract}

}

\maketitle

\IEEEdisplaynontitleabstractindextext

%
\IEEEpeerreviewmaketitle

\IEEEraisesectionheading{\section{Introduction}\label{sec:introduction}}

%
%
%
%

\IEEEPARstart{R}{ecently}, instruction tuning has shown effectiveness in bridging the gap between diverse vision-language tasks and creating general-purpose Multimodal Large Language Models (MLLMs) with broad capabilities \cite{dai2023instructblip,liu2023visual}. MLLMs are always expected to learn new vision-language tasks to support new functionalities in real-world applications. However, existing MLLMs are static and cannot meet continuously emerging new requirements. To avoid the massive cost of retraining MLLMs, we can formulate this scenario into the paradigm of Multimodal Continual Instruction Tuning (MCIT). In MCIT, we aim to instruction-tune MLLMs for new multimodal tasks incrementally while maintaining superior performance on learned tasks.

For example, we start from InstructBLIP \cite{dai2023instructblip}, an MLLM instruction-tuned on 13 datasets. Then, we instruction-tune InstructBLIP incrementally on Flickr30k \cite{young2014image}, VizWiz \cite{gurari2018vizwiz}, TextVQA \cite{singh2019towards}, and GQA \cite{hudson2019gqa}. Ultimately, we expect the final model to perform well on both 4 downstream and 13 pre-trained tasks. Therefore, MCIT is an ideal scenario that equips MLLMs with new skills incrementally without retraining on all datasets.

\begin{figure}[!t]
    \centering
    \includegraphics[width=0.95\linewidth]{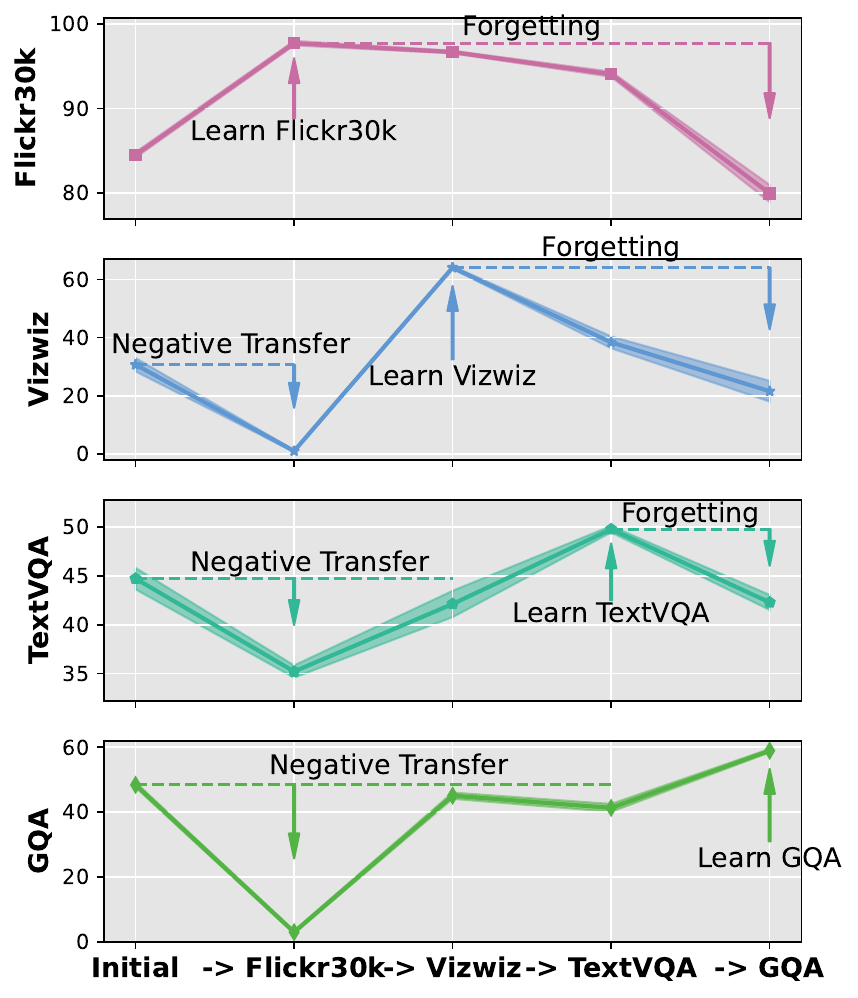}
    \caption{Catastrophic forgetting and negative forward transfer in multimodal continual instruction-tuning. InstructBLIP \cite{dai2023instructblip} is sequentially instruction-tuned on Flickr30k, VizWiz, TextVQA, and GQA.}
    \label{fig:motivation}
\end{figure}

However, MCIT is particularly challenging. Unlike typical continual learning scenarios, the tasks in MCIT are diverse in input and output formats. For example, in Flickr30k, the model outputs a detailed description given an image. In GQA, the model outputs a short answer given an image and a question. In TextVQA, the model outputs a short answer given an image, a question, and OCR (Optical Character Recognition) tokens. Therefore, MLLMs are required to generate various responses according to the task instruction.

In this paper, we observe that diversity in input and output formats exacerbates the interference between tasks. For example, in Figure~\ref{tab:case_study}, models tend to output \emph{shorter} responses in image captioning tasks after training on QA tasks. In this case, MCIT faces two crucial challenges: \emph{catastrophic forgetting} and \emph{negative forward transfer}. By catastrophic forgetting, we mean models forget old knowledge when learning new tasks. For instance, in Figure~\ref{fig:motivation}, the performance of Flickr30k (the topmost subfigure) drops significantly when the model is instruction-tuned on new tasks. By negative forward transfer, we mean that the performance of unseen tasks is degraded when learning new tasks. For instance, after instruction-tuning InstructBLIP on Flickr30K, the zero-shot performance on the unseen tasks (i.e., VizWiz, TextVQA, and GQA) is negatively impacted. Addressing negative forward transfer is essential because it hinders models from adapting to future tasks. Specifically, GQA's performance drops from 59.19 to 58.92 after sequential training on Flickr30k, VizWiz, and TextVQA compared to direct training on GQA.

Before introducing our methods, let us take a closer look at catastrophic forgetting and negative forward transfer. The old knowledge should be preserved as much as possible to achieve anti-forgetting. However, in MCIT, not all old knowledge benefits new tasks. For instance, OCR-VQA \cite{mishra2019ocr} requires models to recognize the text in images locally, while Flickr30k requires understanding the image globally. As shown empirically by \cite{he2023continual}, regularizing old models leads to poor performance on new tasks. In other words, only mitigating catastrophic forgetting may lead to negative forward transfer. Therefore, we need to identify what to preserve and update for each task to avoid both issues.

We perform SVD \cite{golub1971singular} on input embeddings to gain more insight. We discover a large \emph{discrepancy} between input embeddings from different tasks. More importantly, the rank of input embeddings decreases when directly instruction-tuning MLLMs on each new task. However, it increases under the continual learning paradigm. This phenomenon indicates that the discrepancy results in models extracting irrelevant input information for old tasks when adapting to new tasks, which leads to catastrophic forgetting. Moreover, negative forward transfer happens because adapting to the current task results in the model extracting irrelevant information for unseen tasks.

Motivated by these observations, we propose Prompt Tuning with Positive Forward Transfer (henceforth Fwd-Prompt), which achieves anti-forgetting and positive forward transfer. Specifically, Fwd-Prompt proposes selecting prompts according to visual and textual features. On the one hand, Fwd-Prompt achieves anti-forgetting by \emph{allocating different subspaces for each task} and projecting prompt gradient to the residual space. On the other hand, Fwd-Prompt achieves positive forward transfer by \emph{reusing pre-trained knowledge} and projecting prompt gradient to the pre-trained space. Finally, the experiment indicates that Fwd-Prompt achieves state-of-the-art (SOTA) performance on MCIT. In summary, our contribution is threefold:
\begin{itemize}
    \item We identify that the key challenge of MCIT is to achieve both anti-forgetting and positive forward transfer.
    \item We analyze the space of input embeddings and reveal that the discrepancy in input embeddings leads to catastrophic forgetting and negative forward transfer.
    \item To address these two issues, we propose Fwd-Prompt for MCIT, which outperforms SOTA methods by a large margin while requiring fewer trainable parameters and no rehearsal data.
\end{itemize}

\section{Related Work}

\section{Multimodal Large Language Models}
The remarkable capabilities of Multimodal Large Language Models (MLLMs) \cite{openai2023gpt4} have captured widespread attention within the artificial intelligence community, showcasing their proficiency in comprehending, reasoning, and generating human language. 
Despite the considerable progress achieved, these methods face challenges when confronted with more complex multimodal information \cite{mishra2019ocr,sidorov2020textcaps,hudson2019gqa}.
Instruction tuning originated from natural language processing and has become a popular strategy for aligning MLLMs with human intent \cite{touvron2023llama,chiang2023vicuna}.
\cite{dai2023instructblip} argue that instruction-tuned MLLMs generalize to unseen tasks better than multitask learning without instructions.
They further instruction-tune BLIP2 on 13 high-quality vision-language datasets and proposed InstructBILP.
Motivated by InstructBLIP, recent studies \cite{panagopoulou2023x,gong2023multimodal,wu2023next} further extend instruction-tuning to video, audio, and point cloud modalities.
In this study, we focus on vision-language tasks because of their popularity.

\subsection{Continual Learning}
Although a vast number of continual learning methods have been proposed in both computer vision and natural language processing, most of them are not applied to MCIT because they are either inapplicable to multimodal tasks \cite{sun2019lamol} or computationally expensive for MLLMs \cite{buzzega2020dark}. Recently, prompt-based methods such as L2P and CODA-Prompt \cite{wang2022learning,smith2023coda,wang2022dualprompt,qiao2023prompt} have achieved superior performance in continual image classification. These methods harness the power of pre-trained knowledge by freezing pre-trained models and learning only soft prompts for continual learning. Inspired by their success in mitigating forgetting, we propose prompt selection and subspace allocation mechanisms to encourage positive forward transfer in MCIT.

The proposed Fwd-Prompt differs from L2P \cite{wang2022learning} in the following aspects:
1) Fwd-Prompt avoids forgetting and negative transfer by allocating different subspaces to each task (Section \ref{sec:gradient-project});
2) Fwd-Prompt learns a multimodal prompt pool, enabling the incorporation of information from multiple modalities jointly instead of separately (Section \ref{sec:prompt-pool});
3) L2P is only applicable for classification tasks because it requires a classifier, while Fwd-Prompt can be applied to generation tasks, offering a broader range of applications.

The proposed Fwd-Prompt is different from CODA-Prompt \cite{smith2023coda} as follows: CODA-Prompt encourages orthogonality by intuitively adding constraints to the loss function, where orthogonality between tasks is not guaranteed. In contrast, Fwd-Prompt is theoretically grounded and guarantees orthogonality by allocating different subspaces.

More recently, \cite{qiao2023prompt} proposes incorporating the idea of gradient projection \cite{saha2021gradient} into prompt tuning for continual learning. Fwd-Prompt is partially inspired by gradient projection in \cite{saha2021gradient,qiao2023prompt}. Apart from the multimodal prompt pool, Fwd-Prompt designs an algorithm to identify pre-trained space, residual space, and conflicting spaces for each task to achieve both forward knowledge transfer and preserving old knowledge, whereas existing methods \cite{saha2021gradient,qiao2023prompt} achieve only anti-forgetting.

\subsection{Continual Learning with MLLMs}
Continual learning with MLLMs is an emerging research field. Existing research can be divided into two categories: continual learning of visual question answering (VQA) and continual vision-language pretraining.
(1) Continual learning of visual question answering (VQA): 
\cite{srinivasan2022climb} proposes CLiMB, a continual learning benchmark for vision-language, vision-only, and language-only classification tasks. CL-CrossVQA \cite{zhang2022cl} and VQACL \cite{zhang2023vqacl} are two benchmarks for cross-domain VQA and VQA with skill-concept compositions, respectively. \cite{qian2023decouple} designs particular strategies for interactions between modalities.
(2) Continual vision-language pretraining: 
These methods \cite{ni2023continual,zheng2023preventing,zhu2023ctp} aim at aligning vision and language features. Unlike continual VQA, the models are trained on image-text pairs and typically evaluated on retrieval or zero-shot classification tasks.

Unlike the studies above, we explore the continual learning scenario of instruction tuning. \emph{There are two key differences between MCIT and the aforementioned continual scenarios.}
1) The MLLM outputs natural language directly and requires no additional classifiers.
2) MCIT allows MLLMs to learn diverse tasks beyond VQA, including image captioning, image captioning with OCR tokens, and VQA with OCR tokens.

To our knowledge, we are the first to identify forgetting and negative forward transfer in MCIT, and Fwd-Prompt is the first method in MCIT that achieves both anti-forgetting and positive forward transfer. Fwd-Prompt is applicable to state-of-the-art MLLMs such as InstructBLIP and BLIP-2, while existing methods such as L2P and CODA-Prompt are not.

\section{Preliminaries}
\subsection{Prompt-Based Continual Learning}
Prompt-based continual learning is first proposed by \cite{wang2022learning}.
Typically, there is a set of key-value pairs called prompt pool $\mathcal{P}=\{\boldsymbol{k_j},\boldsymbol{p_j}\}_j^{M}$, where $M$ is the number of total key-value pairs, $\boldsymbol{k_j}, \boldsymbol{p_j}\in \mathbb{R}^{d}$ represent the $j$-th key and value, respectively.
For an input image $\boldsymbol{x_i}$, its query feature is computed as $\boldsymbol{q_i}=f^{(img)}(\boldsymbol{x_i})$, where $f^{(img)}(\cdot)$ is usually a pre-trained and frozen Vision Transformer (ViT).
Then, we select the top-$n_p$ prompts according to the cosine similarity between $\boldsymbol{q_i}$ and all keys in the prompt pool.
Finally, we concatenate the selected prompts to the input embeddings of $\boldsymbol{x_i}$ and feed them into the backbone model such as ViT.

\subsection{Singular Value Decomposition}
Singular Value Decomposition (SVD) \cite{golub1971singular} is useful in linear algebra.
Typically, the singular value decomposition of an matrix $\boldsymbol{A}\in \mathbb{R}^{M \times N}$ is $\boldsymbol{A}= \boldsymbol{U}\boldsymbol{\Sigma}\boldsymbol{V}^{T}$, where $\boldsymbol{U}\in \mathbb{R}^{M\times M}$ and $\boldsymbol{V}\in \mathbb{R}^{N\times N}$ are two orthogonal matrices, and $\boldsymbol{\Sigma}\in \mathbb{R}^{M\times N}$ is a rectangular diagonal matrix.
The columns of $\boldsymbol{U}$ and $\boldsymbol{V}$ form two sets of orthonormal bases called left-singular vectors $\{\boldsymbol{u_1},\boldsymbol{u_2},\cdots,\boldsymbol{u_M}\}$ and right-singular vectors $\{\boldsymbol{v_1},\boldsymbol{v_2},\cdots,\boldsymbol{v_N}\}$.
The main diagonal of $\boldsymbol{\Sigma}$ is the singular values $\sigma_1,\sigma_2, \cdots, \sigma_r$, where $r<\min\{M, N\}$ is the rank of $\boldsymbol{A}$ and the singular values are sorted in a descending order.

We can regard $\boldsymbol{A}$ as a transformation from $\mathbb{R}^{N}$ to $\mathbb{R}^{M}$.
Then, $\boldsymbol{v_1}$ is the unit vector that is maximally stretched by $\boldsymbol{A}$, i.e., $\arg\max_{x:||\boldsymbol{x}||_2=1}||\boldsymbol{A}\boldsymbol{x}||_2 = \boldsymbol{v_1}$.
We provide a brief proof as follows.
First, we convert the objective as follows: $\arg\max_{\boldsymbol{x}:||\boldsymbol{x}||_2=1}||\boldsymbol{A}\boldsymbol{x}||_2=\arg\max_{\boldsymbol{x}:||\boldsymbol{x}||_2=1}||\boldsymbol{A}\boldsymbol{x}||_2^{2}=\arg\max_{\boldsymbol{x}:||\boldsymbol{x}||_2=1}\boldsymbol{x}^{T}\boldsymbol{A}^{T}\boldsymbol{A}\boldsymbol{x}$.
Since the eigenvalue expression of $\boldsymbol{A}^{T}\boldsymbol{A}$ is $(\boldsymbol{A}^{T}\boldsymbol{A})\boldsymbol{x}=\lambda \boldsymbol{x}$, we have $\boldsymbol{x}^{T}\boldsymbol{A}^{T}\boldsymbol{A}\boldsymbol{x}=\lambda$.
Then, the objective is converted to choose $x$ as the eigenvector with the largest eigenvalue. 
Because $\boldsymbol{A}^{T}\boldsymbol{A} = \boldsymbol{V} \boldsymbol{\Sigma}^2 \boldsymbol{V}^{T}$ and $\boldsymbol{V}$ is given by the eignedecomposition of $\boldsymbol{A}^{T}\boldsymbol{A}$, $\boldsymbol{v_1}$ is exactly the unit vector that maximizes the objective $||\boldsymbol{A}\boldsymbol{x}||_2$.
Similarly, we can prove that $\boldsymbol{v_N}$ is the unit vector minimally stretched by $\boldsymbol{A}$.

Furthermore, we can divide $\boldsymbol{V}$ into two orthogonal subspace $\boldsymbol{V_{core}}=[\boldsymbol{v_1},\boldsymbol{v_2},\cdots,\boldsymbol{v_K}]$ and $\boldsymbol{V_{res}}=[\boldsymbol{v_{K+1}},\boldsymbol{v_{K+2}},\cdots,\boldsymbol{v_N}]$ by selecting the first $K$ and the remaining eigenvectors, denoted as the \emph{core space} and the \emph{residual space} respectively. 
Since $\boldsymbol{V}\boldsymbol{V}^{T}=I$ and $I$ represents the identity matrix, we have $\boldsymbol{V_{core}}\boldsymbol{V_{core}}^{T}+\boldsymbol{V_{res}}\boldsymbol{V_{res}}^{T}=\boldsymbol{I}$.
For an arbitrary vector $\boldsymbol{x}$, we can decompose it into the core and residual space as $\boldsymbol{x} = \boldsymbol{x_{core}}+\boldsymbol{x_{res}}$, where $\boldsymbol{x_{core}}= \boldsymbol{V_{core}}\boldsymbol{V_{core}}^{T}\boldsymbol{x}$ and $\boldsymbol{x_{res}} = \boldsymbol{V_{res}}\boldsymbol{V_{res}}^{T}\boldsymbol{x}$.
Then, the output vector of $\boldsymbol{x_{core}}$ (i.e.,$\boldsymbol{A}\boldsymbol{x_{core}}$) has a large norm, while the output vector of $\boldsymbol{x_{res}}$ has a small norm.
This way, we can control the impact of $\boldsymbol{x}$ on the output vector by mapping it into different subspaces before multiplying with $\boldsymbol{A}$.

\begin{figure}[!t]
    \centering
    
    \subfloat[]{
        \includegraphics[width=0.49\linewidth]{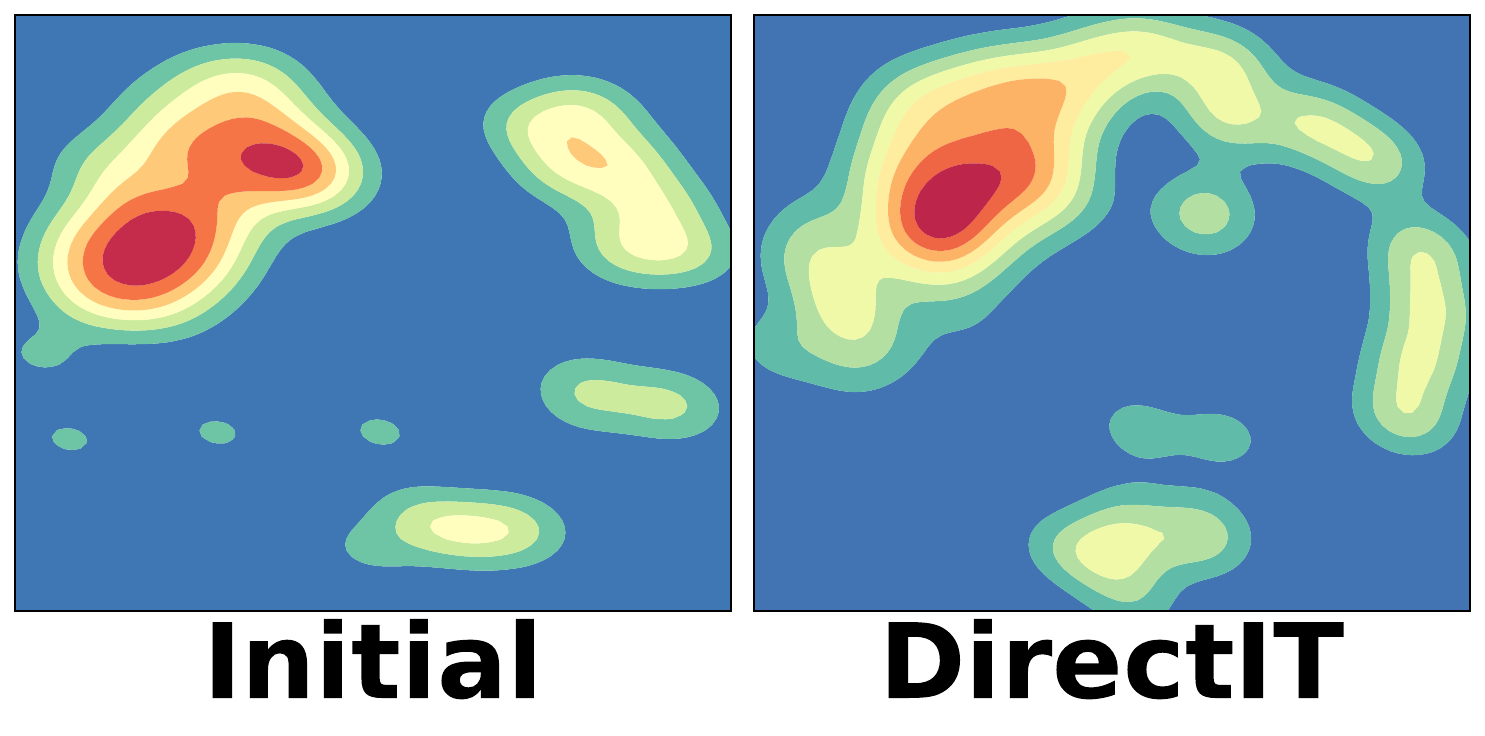}
    }
    \subfloat[]{
        \includegraphics[width=0.49\linewidth]{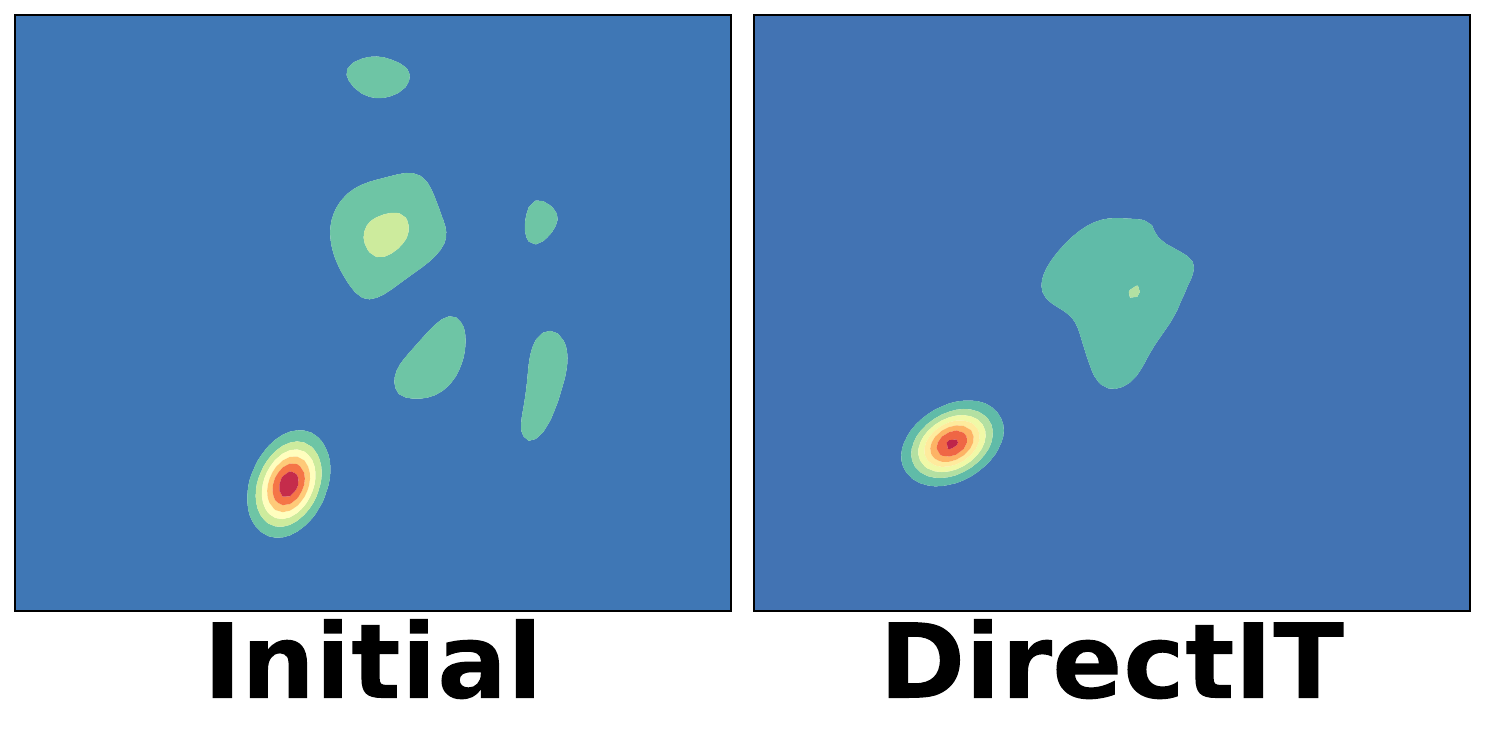}
    }
    
    \subfloat[]{
        \includegraphics[width=0.49\linewidth]{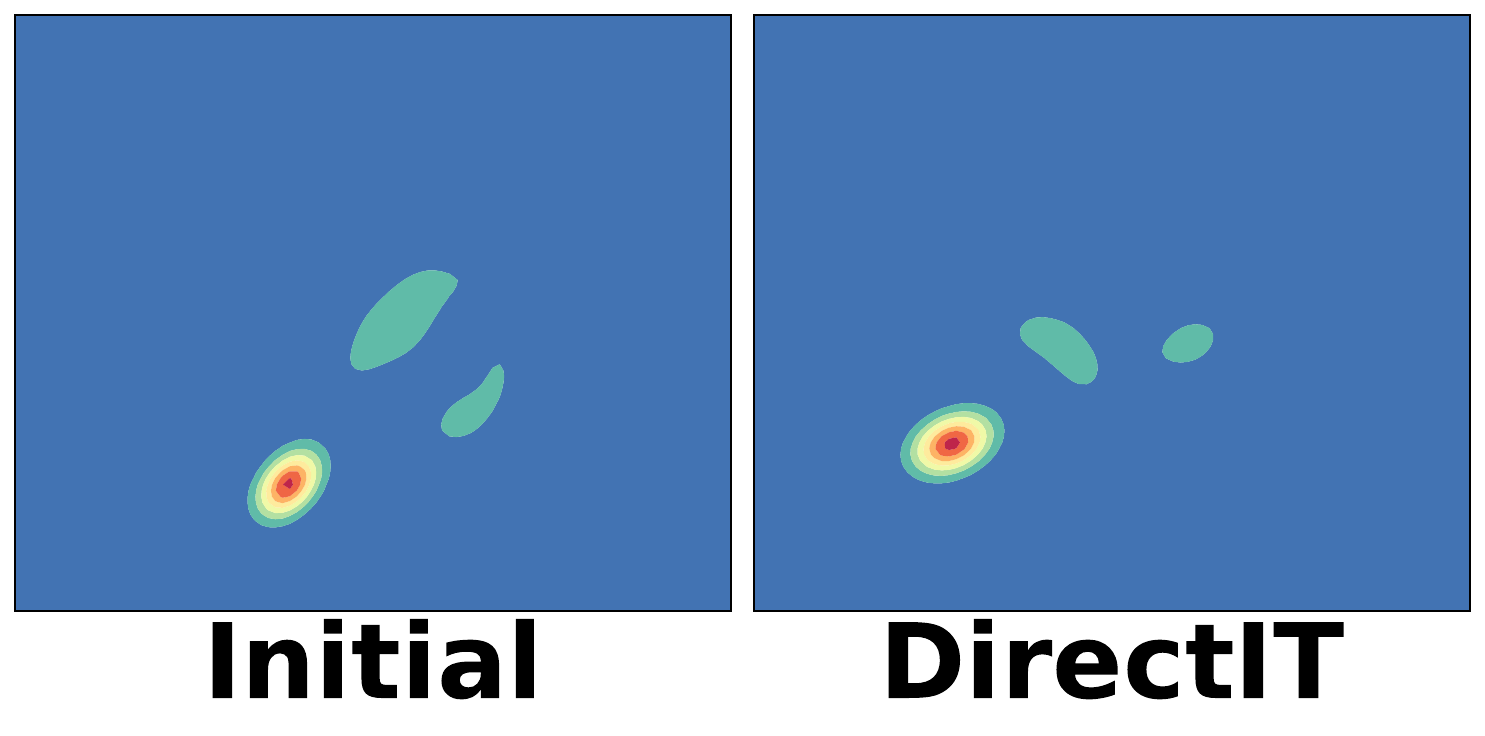}
    }
    \subfloat[]{
        \includegraphics[width=0.49\linewidth]{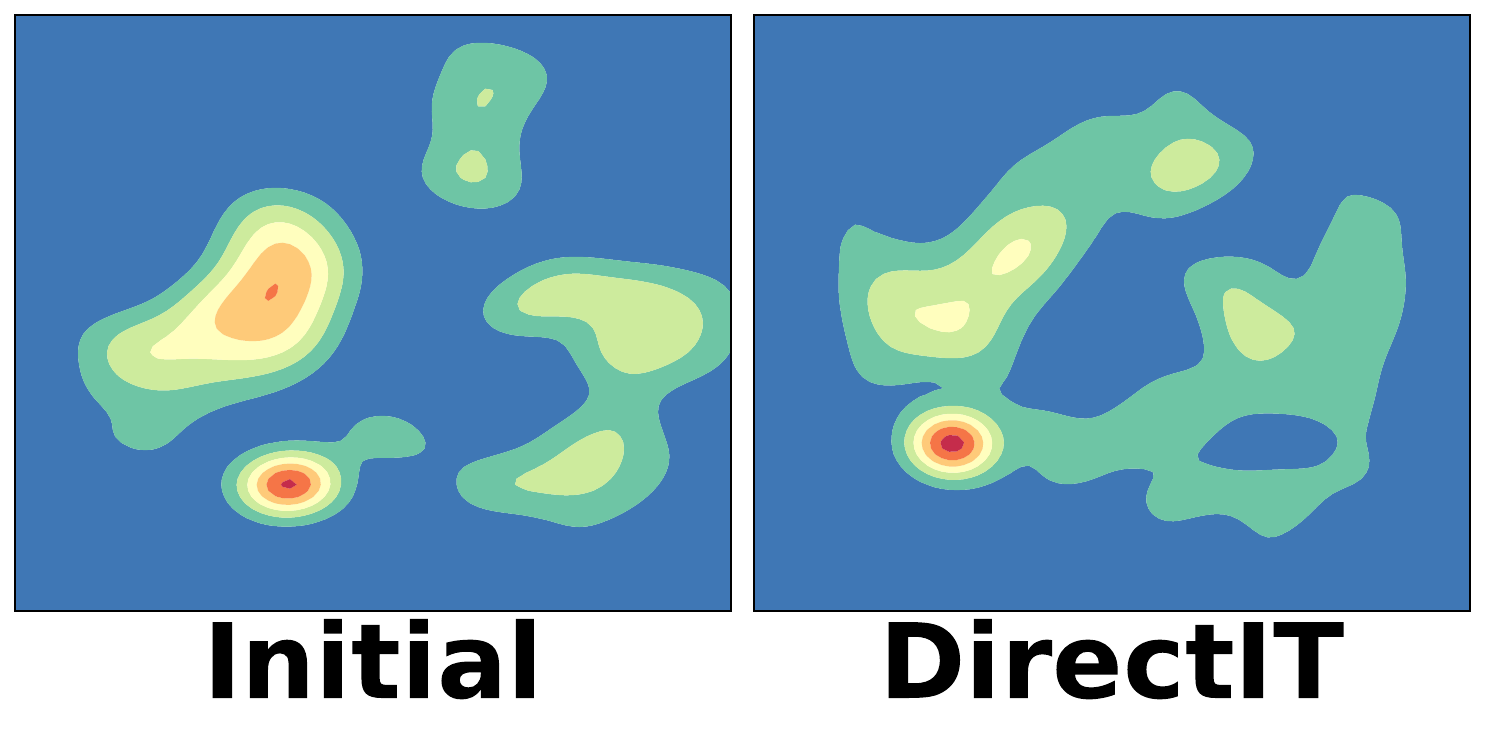}
    }

    \caption{The contour of the distribution of input embeddings from four different tasks. ``Initial'' and ``DirectIT'' represent the input embeddings of InstructBLIP before and after instruction tuning on each task, respectively. (a) Flickr30k. (b) VizWiz. (c) TextVQA. (d) GQA.}
    \label{fig:visualization_contour_diff_task}
\end{figure}

\begin{figure}[!t]
    \centering
    \includegraphics[width=0.99\linewidth]{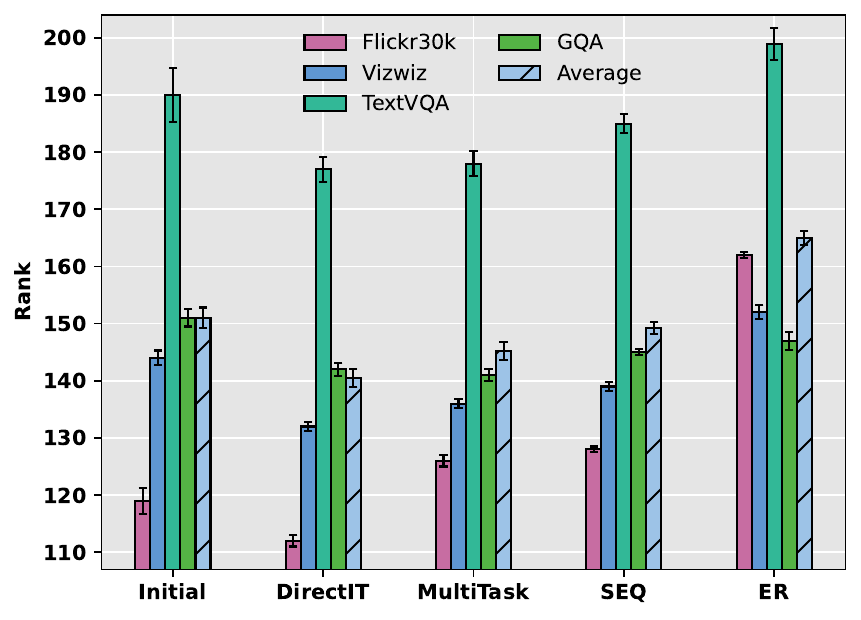}
    \caption{The rank of input embeddings in five scenarios. ``Initial'': load InstructBLIP without training; ``DirectIT'': train InstructBLIP on the four datasets respectively; ``MultiTask'': train InstructBLIP on all four datasets jointly; ``SEQ'': train InstructBLIP on the four datasets sequentially; ``ER'': store 1 \% of old instances and train InstructBLIP sequentially using the experience replay strategy. 
    ``Initial'' is a starting point of MCIT.
    ``DirectIT'' and ``MultiTask'' are both the upper bound of MCIT and ``SEQ'' is the lower bound.
    ``ER'' is a popular technique for alleviating forgetting.}
    \label{fig:rank_diff_task}
\end{figure}

\section{Methodology}
SVD analyzes how many directions (i.e., rank) are needed to represent each task's input. In Section~\ref{sec:a_closer_look_at_forgetting}, we qualitatively and quantitatively analyze the input embedding space in continual learning to gain insights into forgetting and forward transfer. Specifically, we first observe the diversity of the tasks in MCIT in Figure \ref{fig:visualization_contour_diff_task} and validate our assumption about forgetting and negative transfer through the \emph{``rank'' experiment} in Figure \ref{fig:rank_diff_task}. Motivated by these findings, we propose Fwd-Prompt to mitigate the interference between tasks from the \emph{``rank'' perspective} to achieve both anti-forgetting and positive transfer. The description of Fwd-Prompt is provided in Sections~\ref{sec:prompt-pool}, \ref{sec:prompt-anti-forgetting}, and \ref{sec:gradient-project}.

\subsection{A Closer Look at Forgetting and Negative Forward Transfer}
\label{sec:a_closer_look_at_forgetting}
MCIT is challenging due to the considerable discrepancy in input and output formats across tasks. To understand how this affects model performance, we visualize the input embeddings of each task before (Initial) and after (DirectIT) instruction-tuning InstructBLIP on each task. InstructBLIP only trains Q-Former to align the visual and textual features while keeping the ViT and LLM frozen. Here, input embeddings refer to the visual and textual embeddings extracted by Q-Former to feed into the LLM.

Figure~\ref{fig:visualization_contour_diff_task} shows that the distribution of input embeddings has a large discrepancy across tasks. Specifically, we find that Flickr30k has a broader distribution than the other three VQA-based tasks, suggesting that image captioning requires detailed information from the input image. This explains why the performance of VQA-based tasks degrades when learning Flickr30k (Figure~\ref{fig:motivation}), as the differences in information requirements for each task are significant. Figure~\ref{fig:visualization_contour_diff_task} also shows that the distribution of ``Initial'' and ``DirectIT'' is similar, indicating that the required information remains relatively stable during instruction-tuning.

To quantitatively analyze the changes in input embeddings during continual learning, we estimate the rank of input embeddings with a threshold of $\epsilon=0.99$. Figure~\ref{fig:rank_diff_task} shows that ``DirectIT'' and ``MultiTask'' have a lower rank than ``Initial'', indicating that all tasks can be jointly learned without introducing additional input information. However, under continual learning scenarios (``SEQ'' and ``ER''), the rank of old tasks (Flickr30k, VizWiz, and TextVQA) increases significantly. This indicates that the model learns to extract irrelevant input information for old tasks when adapting to new tasks, leading to catastrophic forgetting. From this perspective, negative forward transfer occurs because adapting to the current task results in the model extracting irrelevant information for unseen tasks.

\begin{figure}[!t]
    \centering
    \includegraphics[width=0.99\linewidth]{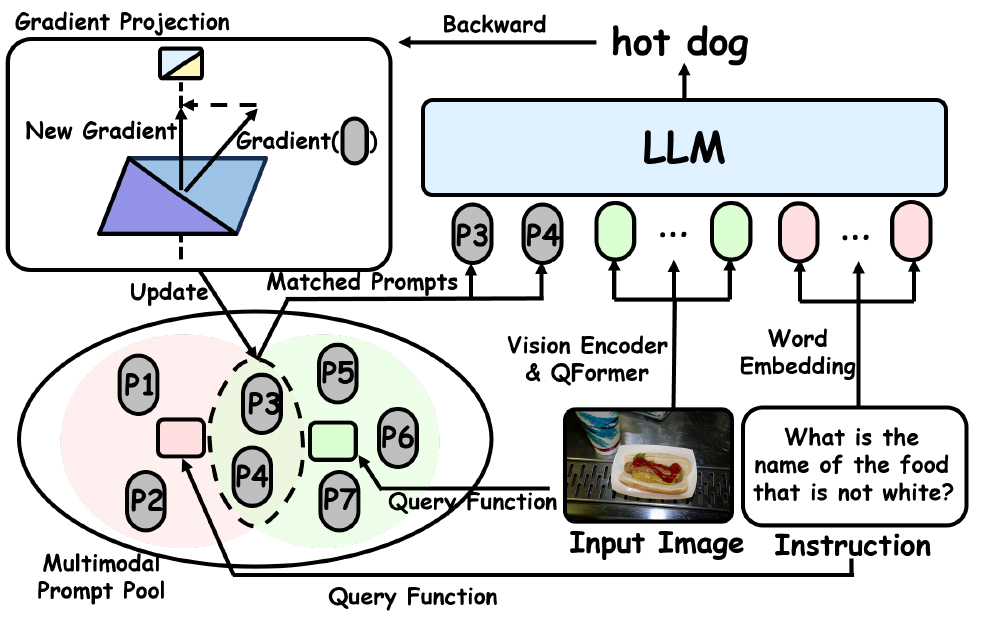}
    \caption{An overview of Fwd-Prompt. The multimodal prompt pool and gradient projection details are in Sections~\ref{sec:prompt-pool} and \ref{sec:gradient-project}, respectively. We provide multimodal prompt pool and gradient projection illustrations in Figures \ref{fig:method_joint_similarity}, \ref{fig:method_gradient_projection}, respectively.}
    \label{fig:method_overview}
\end{figure}

\begin{figure}[!t]
    \centering
    \includegraphics[width=0.99\linewidth]{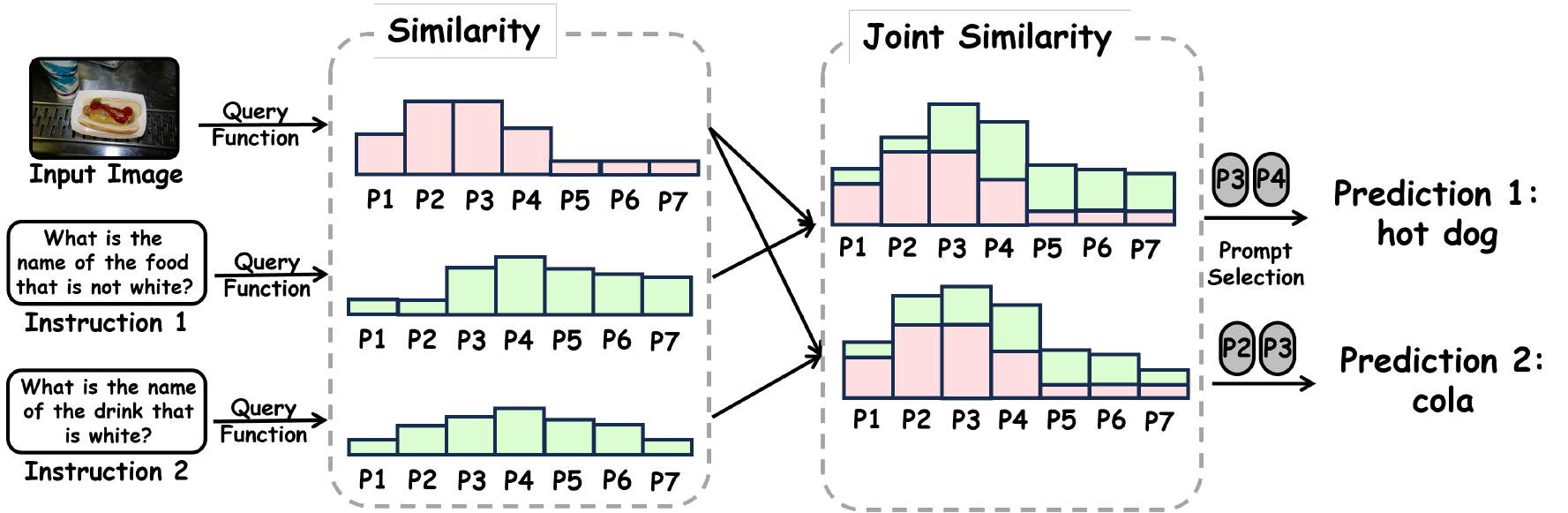}
    \caption{An illustration of learning a multimodal prompt pool with joint similarity. The key intuition for building a multimodal prompt pool is that both image and text instruction should determine each prompt. For example, we expect the model to select different prompts when different text instructions are provided for the same input image.}
    \label{fig:method_joint_similarity}
\end{figure}

\subsection{Learning a Multimodal Prompt Pool with Joint Similarity}
\label{sec:prompt-pool}
We provide an overview of Fwd-Prompt in Figure~\ref{fig:method_overview}. This subsection describes how to design a multimodal prompt pool for multimodal tasks.

Intuitively, it is easier for the model to output the correct answer if task-specific descriptions such as ``This task requires reading the text on books'' or instance-specific descriptions such as ``This photo was taken at a fast food restaurant'' are provided. Therefore, we follow L2P \cite{wang2022learning} and design a prompt pool where soft prompts are selected dynamically according to each instance.

In MCIT, the output is determined by both the input image and instruction. For instance, given the same input image in Figure~\ref{fig:method_joint_similarity}, the expected output would be ``Cola'' if the instruction is changed to ``What is the name of the drink on the upper left corner?''. Therefore, each prompt should be determined by both image and instruction. To achieve this, we design two keys for each prompt, and the prompt pool is defined as $\mathcal{P}=\{\boldsymbol{k^{(img)}_j}, \boldsymbol{k^{(text)}_j},\boldsymbol{p_j}\}_j^{M}$, where $M$ is the number of total key-value pairs, $\boldsymbol{p_j}\in \mathbb{R}^{d}$ represent the $j$-th value, $\boldsymbol{k^{(img)}_j}\in \mathbb{R}^{d_i}$ and $\boldsymbol{k^{(text)}_j}\in \mathbb{R}^{d_t}$ represent the key for image and instruction respectively. $d$, $d_i$, and $d_t$ are the dimensions of input embeddings and visual and textual features, respectively. For the $i$-th instance containing an image $\boldsymbol{x_i^{(img)}}$ and instruction $\boldsymbol{x_i^{(text)}}$, its similarity with the $j$-th prompt is calculated as
\begin{equation}
     \phi_{i,j} = \phi(\boldsymbol{q_i^{(img)}},\boldsymbol{k^{(img)}_j}) + \phi(\boldsymbol{q_i^{(text)}},\boldsymbol{k^{(text)}_j}),
     \label{eq:similarity}
\end{equation}
where $\boldsymbol{q_i^{(img)}}$ is the query of $\boldsymbol{x_i^{(img)}}$ and is obtained by averaging the features from a frozen vision encoder such as ViT-g/14 \cite{fang2023eva} used in InstructBLIP. Similarly, $\boldsymbol{q_i^{(text)}}$ is the query of $\boldsymbol{x_i^{(text)}}$ and is obtained by averaging the input embeddings of an LLM such as Flan-T5 \cite{chung2022scaling} used in InstructBLIP. $\phi(\boldsymbol{q_i^{(img)}},\boldsymbol{k^{(img)}_j})$ is computed as the cosine similarity between $\boldsymbol{q_i^{(img)}}$ and $\boldsymbol{k^{(img)}_j}$ normalized over all keys in the prompt pool with the softmax function. $\phi(\boldsymbol{q_i^{(text)}},\boldsymbol{k^{(text)}_j})$ is computed in a similar way.

\subsection{Anti-Forgetting for Prompt Tuning}
\label{sec:prompt-anti-forgetting}
This section describes how to achieve anti-forgetting when utilizing prompt tuning for continual learning.

To achieve the goal of anti-forgetting, we expect the model to output the same response given the same image and instruction when soft prompts are updated. The derivation is inspired by \cite{qiao2023prompt}. Formally, we have
\begin{equation}
    f_{LLM}([\boldsymbol{p_{t+1}^T};\boldsymbol{e^{(img)}_t};\boldsymbol{e^{(text)}_t}]) = f_{LLM}([\boldsymbol{p_{t}^T};\boldsymbol{e^{(img)}_t};\boldsymbol{e^{(text)}_t}]),
    \label{eq:anti-forgetting-objective}
\end{equation}
where $f_{LLM}$ represents the large language model, frozen during continual learning. $\boldsymbol{p_{t+1}}, \boldsymbol{p_{t}} \in \mathbb{R}^{d}$ are the prompts before and after the update of task $t$. We consider only one prompt for demonstration purposes. $\boldsymbol{e^{(img)}_t} \in \mathbb{R}^{n_i \times d}$, $\boldsymbol{e^{(text)}_t} \in \mathbb{R}^{n_t \times d}$ represent the $n_i$, $n_t$ input embedding tokens corresponding to the input image and instruction from task $t$. $[\boldsymbol{p_{t+1}};\boldsymbol{e^{(img)}_t};\boldsymbol{e^{(text)}_t}]$ indicates that the prompt is concatenated with the input along the sequence dimension. We denote the concatenated input $[\boldsymbol{p_{t+1}};\boldsymbol{e^{(img)}_t};\boldsymbol{e^{(text)}_t}]$ and $[\boldsymbol{p_{t}};\boldsymbol{e^{(img)}_t};\boldsymbol{e^{(text)}_t}]$ with a prompt token from task $t$ and task $t+1$ as $\boldsymbol{Z_{t+1}}\in \mathbb{R}^{(1+n_i+n_t)\times d}$ and $\boldsymbol{Z_{t}}\in \mathbb{R}^{(1+n_i+n_t)\times d}$, respectively.

The input embedding tokens primarily interact with each other in the self-attention layer of LLMs. Therefore, we simplify the objective in Equation \ref{eq:anti-forgetting-objective} as follows:
\begin{equation}
    Attention(\boldsymbol{Z_{t+1}}) =  Attention(\boldsymbol{Z_{t}})
    \label{eq:ojective_attention}
\end{equation}
$\boldsymbol{A_{t+1}}$ and $\boldsymbol{A_{t}}$ are the attention matrices of input $\boldsymbol{Z_{t+1}}$ and $\boldsymbol{Z_{t}}$. Then, we expand the attention matrix as follows:
\begin{align}
    Attention(\boldsymbol{Z_{t}}) &= \text{softmax}\left(\frac{\boldsymbol{Z_t} \boldsymbol{W_Q} (\boldsymbol{Z_t} \boldsymbol{W_K})^T}{\sqrt{d_h}}\right) \\
                     &= \text{softmax}\left(\frac{\boldsymbol{Z_t} \boldsymbol{W_Q} \boldsymbol{W_K}^T \boldsymbol{Z_t}^T }{\sqrt{d_h}}\right) 
                     \label{eq:softmax_attention}
\end{align}
where $\boldsymbol{W_Q} \in \mathbb{R}^{d\times d_h}$ and $\boldsymbol{W_K} \in \mathbb{R}^{d\times d_h}$ are the projection matrices of query and key, $\text{softmax}$ represents the row-wise softmax function, and $d_h$ is the hidden dimension. In Equation \ref{eq:softmax_attention}, $\boldsymbol{W_Q}$ and $\boldsymbol{W_K}$ are unchanged during training since LLMs are frozen. To further simplify the objective, we follow \cite{boix2023transformers} and assume the query and key matrices $\boldsymbol{W_Q}$ and $\boldsymbol{W_K}$ are diagonal weight matrices and $d_h = d$:
\begin{equation}
    \boldsymbol{W_Q} \boldsymbol{W_K}^T = \text{diag}(\boldsymbol{w_Q}) \text{diag}(\boldsymbol{w_K})^T,
\end{equation}
where $\boldsymbol{w_Q}, \boldsymbol{w_K} \in \mathbb{R}^{d}$ are the diagonal entries of $\boldsymbol{W_Q}$ and $\boldsymbol{W_K}$. Then, the objective in Equation \ref{eq:ojective_attention} can be simplified as follows:
\begin{equation}
    \boldsymbol{Z_{t+1}} \boldsymbol{Z_{t+1}^T} = \boldsymbol{Z_{t}} \boldsymbol{Z_{t}^T}
    \label{eq:objective_input_multiply}
\end{equation}
Then, we expand the above equation as follows:
\begin{equation}
    \begin{aligned}
    &
    \begin{bmatrix}
        \boldsymbol{p_{t+1}^T} \\
        \boldsymbol{e^{(img)}_t} \\
        \boldsymbol{e^{(text)}_t}
    \end{bmatrix}
    \begin{bmatrix}
    \boldsymbol{p_{t+1}} & \boldsymbol{e^{(img)}_t}^T & \boldsymbol{e^{(text)}_t}^T
    \end{bmatrix}
    = 
    \\
    &\begin{bmatrix}
        \boldsymbol{p_{t}^T} \\
        \boldsymbol{e^{(img)}_t} \\
        \boldsymbol{e^{(text)}_t}
    \end{bmatrix}
    \begin{bmatrix}
        \boldsymbol{p_{t}} & \boldsymbol{e^{(img)}_t}^T & \boldsymbol{e^{(text)}_t}^T
    \end{bmatrix}
    \end{aligned}
\end{equation}

\begin{equation}
    \begin{aligned}
    &
    \begin{bmatrix}
        \boldsymbol{p_{t+1}^T} \boldsymbol{p_{t+1}} & \boldsymbol{p_{t+1}^T} \boldsymbol{e^{(img)}_t}^T & \boldsymbol{p_{t+1}^T} \boldsymbol{e^{(text)}_t}^T \\
        \boldsymbol{e^{(img)}_t} \boldsymbol{p_{t+1}} & \boldsymbol{e^{(img)}_t} \boldsymbol{e^{(img)}_t}^T & \boldsymbol{e^{(img)}_t} \boldsymbol{e^{(text)}_t}^T \\
        \boldsymbol{e^{(text)}_t} \boldsymbol{p_{t+1}} & \boldsymbol{e^{(text)}_t} \boldsymbol{e^{(img)}_t}^T & \boldsymbol{e^{(text)}_t} \boldsymbol{e^{(text)}_t}^T
    \end{bmatrix}
    = 
    \\
    &\begin{bmatrix}
        \boldsymbol{p_{t}^T} \boldsymbol{p_{t}} & \boldsymbol{p_{t}^T} \boldsymbol{e^{(img)}_t}^T & \boldsymbol{p_{t}^T} \boldsymbol{e^{(text)}_t}^T \\
        \boldsymbol{e^{(img)}_t} \boldsymbol{p_{t}} & \boldsymbol{e^{(img)}_t} \boldsymbol{e^{(img)}_t}^T & \boldsymbol{e^{(img)}_t} \boldsymbol{e^{(text)}_t}^T \\
        \boldsymbol{e^{(text)}_t} \boldsymbol{p_{t}} & \boldsymbol{e^{(text)}_t} \boldsymbol{e^{(img)}_t}^T & \boldsymbol{e^{(text)}_t} \boldsymbol{e^{(text)}_t}^T
    \end{bmatrix}
    \end{aligned}\label{eq:large_matrix_equation}
\end{equation}
By comparing the two matrices element-wise in Equation \ref{eq:large_matrix_equation}, we can simplify it as follows:
\begin{numcases}{}
    \boldsymbol{p_{t+1}^T} \boldsymbol{p_{t+1}} = \boldsymbol{p_t^T} \boldsymbol{p_t} \\
    \boldsymbol{e_{t}^{(img)}} \boldsymbol{p_{t+1}} = \boldsymbol{e_{t}^{(img)}} \boldsymbol{p_{t}} \\
    \boldsymbol{e_{t}^{(text)}} \boldsymbol{p_{t+1}} = \boldsymbol{e_{t}^{(text)}} \boldsymbol{p_{t}} 
\end{numcases}
We denote the difference between $\boldsymbol{p_{t+1}}$ and $\boldsymbol{p_{t}}$ as $\boldsymbol{\Delta p}$, i.e., $\boldsymbol{\Delta p} = \boldsymbol{p_{t+1}}-\boldsymbol{p_{t}}$. Finally, the objective in Equation \ref{eq:anti-forgetting-objective} is summarized as follows:
\begin{numcases}{}
    \boldsymbol{p_{t+1}^T} \boldsymbol{p_{t+1}} = \boldsymbol{p_t^T} \boldsymbol{p_t} \label{eq:anti-forgetting-1} \\
    \boldsymbol{e_{t}^{(img)}} \boldsymbol{\Delta p} = \boldsymbol{0} \label{eq:anti-forgetting-2}\\
    \boldsymbol{e_{t}^{(text)}} \boldsymbol{\Delta p} = \boldsymbol{0} \label{eq:anti-forgetting-3}
\end{numcases}
$\boldsymbol{\Delta p} = \boldsymbol{p_{t+1}} - \boldsymbol{p_{t}}$ can be regarded as the gradient of the prompt. Equation \ref{eq:anti-forgetting-1} indicates that the updated prompt should have the same L2 norm. In practice, the learned prompts have relatively close L2 norms, so our implementation ignores the first constraint. The latter two constraints in Equations \ref{eq:anti-forgetting-2} and \ref{eq:anti-forgetting-3} indicate that the prompt gradient should be orthogonal to the input embeddings of both image and instruction. To satisfy these constraints, we perform SVD on the concatenated input embeddings $\boldsymbol{e_t^{(joint)}} = [\boldsymbol{e_t^{(img)}};\boldsymbol{e_t^{(text)}}] \in \mathbb{R}^{(n_i+n_t) \times d}$ and obtain $d$ right singular vectors $\{\boldsymbol{v_1},\boldsymbol{v_2},\cdots,\boldsymbol{v_d}\}$. Then, we select the first $K$ right singular vectors to form the \emph{core space} of task $t$ that satisfy $||\boldsymbol{\widetilde{e_t^{(joint)}}}|| \geq \epsilon||\boldsymbol{e_t^{(joint)}}||$, where $\boldsymbol{\widetilde{e_t^{(joint)}}}$ is the \emph{K-rank approximation} of $\boldsymbol{e_t^{(joint)}}$, $||\cdot||$ represents the Frobenius norm, and $\epsilon=0.99$. The remaining $d-K$ right singular vectors form the \emph{residual space} of task $t$. After obtaining the prompt gradient $\boldsymbol{\Delta p}$, we project $\boldsymbol{\Delta p}$ to the residual space as follows:
\begin{equation}
    \boldsymbol{\Delta p'} = \boldsymbol{V_{res}}\boldsymbol{V_{res}^T} \boldsymbol{\Delta p},
\end{equation}
where $\boldsymbol{V_{res}} = [\boldsymbol{v_{k+1}},\boldsymbol{v_{k+2}},\cdots,\boldsymbol{v_{d}}] \in \mathbb{R}^{d \times (d-K)}$. By mapping the prompt gradient into the residual subspace, the forgetting of the knowledge of task $t$ is minimized.

\begin{figure*}[!t]
    \centering
    \subfloat[]{
        \includegraphics[width=0.45\linewidth]{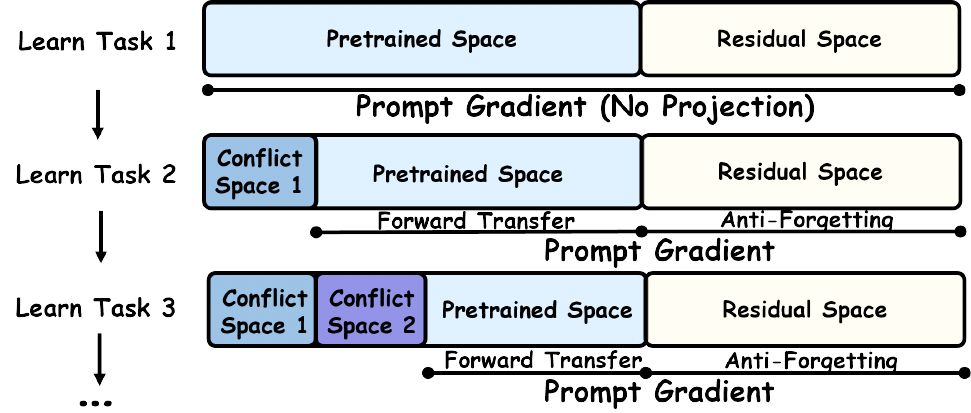}
    }
    \subfloat[]{
        \includegraphics[width=0.45\linewidth]{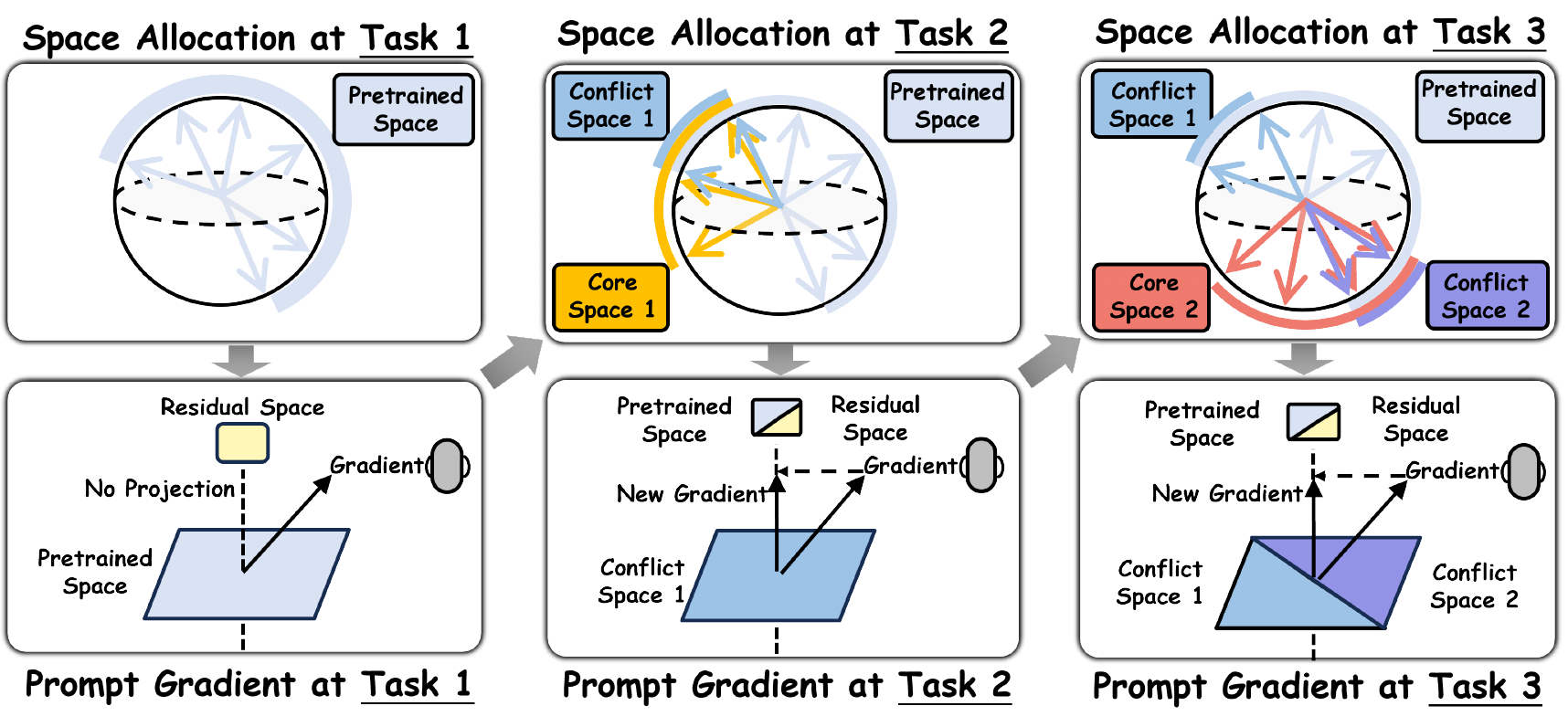}
    }
    \caption{Illustrations for subspace allocation in Fwd-Prompt. (a) An illustration of the subspace allocation when learning each new task. The length indicates the number of directions allocated to the subspace. (b) An illustration of the connection between subspace allocation and prompt gradient projection in Fwd-Prompt. Fwd-Prompt avoids interference between new tasks by allocating different conflicting spaces and projecting gradients to the residual space, thus achieving anti-forgetting.}
    \label{fig:method_gradient_projection}
\end{figure*}

\begin{algorithm}[!t]
    \caption{Fwd-Prompt}
    \label{alg:fwd_prompt}
    \begin{algorithmic}[1]
        \Require Total number of tasks $T$, number of key-value pairs $M$, threshold for $K$-rank approximation $\epsilon$, threshold for finding matching directions $\theta$, and number of selected prompts for input $n_p$
        \State Randomly initialize $M$ key-value pairs $\mathcal{P} = \{\boldsymbol{k_j^{(img)}}, \boldsymbol{k_j^{(text)}}, \boldsymbol{p_j}\}_j^M$
        \State Compute the core space of the pre-trained task $\boldsymbol{V_{pre}}$ using $K$-rank approximation with threshold $\epsilon$
        
        \For{$t \gets 1$ \textbf{to} $T$}
            \State Load the dataset for task $t$
            \For{each instance $(\boldsymbol{x_i^{(img)}}, \boldsymbol{x_i^{(text)}})$ in task $t$}
                \State Compute query features $\boldsymbol{q_i^{(img)}}$ and $\boldsymbol{q_i^{(text)}}$
                \State Calculate similarity $\phi_{i,j}$ between instance and each prompt in $\mathcal{P}$
                \State Select the top-$n_p$ prompts based on similarity
                \State Concatenate the selected prompts with input embeddings
                \State Feed concatenated input into the model and compute the loss
                \State Perform back-propagation and obtain the gradient of the prompt $\boldsymbol{\Delta p}$
                \If{$t = 1$}
                    \State $\boldsymbol{\Delta p'} = \boldsymbol{\Delta p}$
                \Else
                    \State Project the prompt gradient to the orthogonal direction of the conflicting space of tasks $1$ to $t-1$
                    \State $\boldsymbol{\Delta p'} = (\boldsymbol{I} - \boldsymbol{V_{con}^{1,\cdots,t-1}} \boldsymbol{V_{con}^{1,\cdots,t-1}}^T) \boldsymbol{\Delta p}$
                \EndIf
                \State Update prompt with the projected gradient $\boldsymbol{\Delta p'}$
            \EndFor

            \State Perform SVD on input embeddings of task $t$ to identify core space $\boldsymbol{V_{core}^t}$ using $K$-rank approximation with threshold $\epsilon$
            \State Compute conflicting index set $\mathcal{I}^t$:
            \State $\mathcal{I}^t = \{i \mid \exists j, \, |cosine(\boldsymbol{V_{pre}}[:,i], \boldsymbol{V_{core}^t}[:,j])| > \theta\}$
            \State Define the conflicting space of tasks $1$ to $t$:
            \State $\boldsymbol{V_{con}^{1,\cdots,t}} = \{\boldsymbol{V_{pre}}[:,i]\}_{i \in \bigcup_{t' = 1}^{t} \mathcal{I}^{t'}}$
        \EndFor
    \end{algorithmic}
\end{algorithm}

\subsection{Anti-Forgetting and Positive Forward Transfer with Gradient Projection}
\label{sec:gradient-project}

In this subsection, we extend the idea in Section~\ref{sec:prompt-anti-forgetting} to achieve positive forward transfer beyond anti-forgetting. The fundamental intuition is to reuse the pre-trained knowledge during continual learning to avoid extracting irrelevant input information for unseen tasks, as described in Section~\ref{sec:a_closer_look_at_forgetting}. Illustrations for gradient projection in Fwd-Prompt are provided in Figures~\ref{fig:method_gradient_projection}.

Specifically, we regard the pre-trained tasks before continual learning as task $0$ and identify its \emph{pre-trained space} and \emph{residual space} using $K$-rank approximation as described in Section~\ref{sec:prompt-anti-forgetting}. We denote the pre-trained space as $\boldsymbol{V_{pre}} \in \mathbb{R}^{d\times K_0}$ and the residual space as $\boldsymbol{V_{res}} \in \mathbb{R}^{d\times (d-K_0)}$.

When adapting to task $1$, we train the prompt pool without gradient projection. After finishing the training on task $1$, we perform SVD on the input embeddings from task $1$ and obtain the \emph{core space of task $1$ (i.e., core space 1, denoted as $\boldsymbol{V_{core}^{1}}$)} by $K$-rank approximation. Then, we record the indexes of directions in the pre-trained space that conflict with the core space of task $1$. Formally, the conflicting index set $\mathcal{I}^{1}$ for task $1$ is calculated as follows:
\begin{equation}
    \mathcal{I}^{1}=\{i|\exists j, |cosine(\boldsymbol{V_{pre}}[:,i], \boldsymbol{V_{core}^{1}}[:,j])|>\theta \},
\end{equation}
where $\boldsymbol{V_{pre}}[:,i]$ and $\boldsymbol{V_{core}^{1}}[:,j]$ represent the $i$-th and $j$-th column vectors in $\boldsymbol{V_{pre}}$ and $\boldsymbol{V_{core}^{1}}$, respectively. $\theta$ is the threshold for determining if two directions are overlapping. $cosine(\cdot,\cdot)$ computes the cosine similarity between two vectors. Then, we define the \emph{conflicting space of task $\textbf{1}$ (i.e., conflicting space 1)} as follows:
\begin{equation}
    \boldsymbol{V_{con}^{1}} = \{\boldsymbol{V_{pre}}[:,i]\}_{i \in \mathcal{I}^{1}} .
\end{equation}
We note that the core space is defined in terms of each task, while the conflicting space is defined as the overlap between the core space and the pre-trained space. For example, in Figure \ref{fig:method_gradient_projection}, conflicting space 1 is the overlap space between the pre-trained space and core space 1.

When adapting to task $2$, the prompt gradient is projected to the orthogonal direction of $\boldsymbol{V_{con}^{1}}$ to avoid forgetting the knowledge in task $1$:
\begin{equation}
    \boldsymbol{\Delta p'} =  (\boldsymbol{I}-\boldsymbol{V_{con}^{1}}\boldsymbol{{V_{con}^{1}}^T}) \boldsymbol{\Delta p},
\end{equation}
After finishing training on task $2$, we find the conflicting index set $\mathcal{I}^{2}$ for task $2$ in a similar way.

Similarly, when adapting to task $3$, the prompt gradient is computed as follows:
\begin{equation}
    \boldsymbol{V_{con}^{1,2}} = \{\boldsymbol{V_{pre}}[:,i]\}_{i \in \mathcal{I}^{1} \cup \mathcal{I}^{2}} .
    \label{eq:gradient_project_task3_index}
\end{equation}
\begin{equation}
    \boldsymbol{\Delta p'} =  (\boldsymbol{I}-\boldsymbol{V_{con}^{1,2}}\boldsymbol{{V_{con}^{1,2}}^T}) \boldsymbol{\Delta p}.
    \label{eq:gradient_project_task3}
\end{equation}
In Equation \ref{eq:gradient_project_task3_index}, we compute the union of the indexes of conflicting spaces 1 and 2. In Equation \ref{eq:gradient_project_task3}, we project the gradient to the orthogonal directions of conflicting spaces 1 and 2 to minimize the interference to tasks 1 and 2 when learning task 3. We summarize the algorithm in Algorithm \ref{alg:fwd_prompt}.

In summary, Fwd-Prompt avoids interference between new tasks by allocating different conflicting spaces and projecting gradients to the residual space, thus achieving anti-forgetting. Furthermore, Fwd-Prompt reuses pre-trained knowledge by updating prompts in the pre-trained space, which contains the core space of unseen tasks, thus enhancing positive forward transfer.

\begin{table*}[!t]
  \centering
  \caption{The dataset statistics when \emph{InstructBLIP} is used as the backbone MLLM for MCIT.}
  \resizebox{0.9\linewidth}{!}{
    \begin{tabular}{cllllll}
    \toprule
    \textbf{Default Order} & \textbf{Dataset} & \textbf{Task Description} & \textbf{\# Training Instance} & \textbf{\# Dev Instance} & \textbf{Metric} & \textbf{Source} \\
    \midrule
    \multirow{6}[2]{*}{0 } & Caption COCO \cite{chen2015microsoft} & Image Captioning & 82K   & 5K    & CIDEr & \href{https://cocodataset.org/}{Link} \\
          & TextCaps \cite{sidorov2020textcaps} & Image Captioning with OCR tokens & 109K  & 16K   & CIDEr & \href{https://textvqa.org/textcaps/}{Link} \\
          & VQAv2 \cite{goyal2017making} & VQA   & 443K  & 214K  & ACC   & \href{https://visualqa.org/}{Link} \\
          & OKVQA \cite{marino2019ok} & VQA   & 9K    & 5K    & ACC   & \href{https://okvqa.allenai.org/}{Link} \\
          & A-OKVQA \cite{schwenk2022okvqa} & VQA   & 17K   & 1K    & ACC   & \href{https://github.com/allenai/aokvqa}{Link} \\
          & OCR-VQA \cite{mishra2019ocr} & VQA   & 800K  & 100K  & ACC   & \href{https://ocr-vqa.github.io/}{Link} \\
    \midrule
    1     & Flickr30k \cite{young2014image} & Image Captioning & 145K  & 1K    & CIDEr & \href{https://www.kaggle.com/datasets/hsankesara/flickr-image-dataset}{Link} \\
    2     & VizWiz \cite{gurari2018vizwiz} & VQA   & 20K   & 4K    & ACC   & \href{https://vizwiz.org/}{Link} \\
    3     & TextVQA \cite{singh2019towards} & VQA with OCR tokens & 34K   & 5K    & ACC   & \href{https://textvqa.org/dataset/}{Link} \\
    4     & GQA \cite{hudson2019gqa} & VQA   & 94K   & 13K   & ACC   & \href{https://downloads.cs.stanford.edu/nlp/data/gqa/images.zip}{Link} \\
    \bottomrule
    \end{tabular}%
    }
  \label{tab:dataset_statistics_instructblip}%
\end{table*}%

\begin{table*}[!t]
  \centering
  \caption{The dataset statistics when \emph{BLIP2} is used as the backbone MLLM for MCIT.}
  \resizebox{0.9\linewidth}{!}{
    \begin{tabular}{cllllll}
        \toprule
    \textbf{Default Order} & \textbf{Dataset} & \textbf{Task Description} & \textbf{\# Training Set} & \textbf{\# Dev Set} & \textbf{Metric} & \textbf{Source} \\
    \midrule
    1     & Flickr30k \cite{young2014image} & Image Captioning & 145K  & 1K    & CIDEr & \href{https://www.kaggle.com/datasets/hsankesara/flickr-image-dataset}{Link} \\
    2     & TextCaps \cite{sidorov2020textcaps} & Image Captioning with OCR tokens & 549K  & 16K   & CIDEr & \href{https://textvqa.org/textcaps/}{Link} \\
    3     & VQAv2  \cite{goyal2017making} & VQA   & 443K  & 214K  & ACC   & \href{https://visualqa.org/}{Link} \\
    4     & OCR-VQA \cite{mishra2019ocr} & VQA   & 801K  & 100K  & ACC   & \href{https://ocr-vqa.github.io/}{Link} \\
    5     & GQA \cite{schwenk2022okvqa} & VQA   & 94K   & 13K   & ACC   & \href{https://downloads.cs.stanford.edu/nlp/data/gqa/images.zip}{Link} \\
    \bottomrule
    \end{tabular}%
    }
  \label{tab:dataset_statistics_blip2}%
\end{table*}%

\begin{table}[!t]
  \centering
  \caption{The templates for inference of Fwd-Prompt and all compared baseline methods.}
  \resizebox{\linewidth}{!}{
    \begin{tabular}{c|c}
    \toprule
    Datasets & Instruction Templates \\
    \midrule
    \makecell{VQAv2, OKVQA, A-OKVQA, \\OCR-VQA, VizWiz, GQA} & $<$Image$>$ Question: \{\} Short answer: \\
    \midrule
    Flickr30k, Caption COCO, & $<$Image$>$ A short image description: \\
    \midrule
    TextVQA & $<$Image$>$ OCR tokens: \{\}. Question: \{\} Short answer: \\
    \midrule
    TextCaps & $<$Image$>$ OCR tokens: \{\}. A short image description: \\
    \bottomrule
    \end{tabular}%
    }
  \label{tab:prompt-summary}%
\end{table}%

\begin{table}[!t]
  \centering
  \caption{The training hyper-parameters used for training Fwd-Prompt and all compared baseline methods.}
  \resizebox{\linewidth}{!}{
    \begin{tabular}{c|c}
    \toprule
    Hyper-Parameters & Value \\
    \midrule
    max epoch & 1 for GQA and 5 for others \\
    \midrule
    initial lr & 1.00E-05 \\
    \midrule
    warmup lr & 1.00E-08 \\
    \midrule
    warmup steps & 1000 \\
    \midrule
    weight decay & 0.05 \\
    \midrule
    batch size & 32 \\
    \midrule
    num beams & 5 \\
    \midrule
    \multicolumn{1}{c|}{\multirow{2}[4]{*}{output len}} & VQA-based Tasks: max\_len=10, min\_len=1 \\
\cmidrule{2-2}          & Image Captioning Tasks: max\_len = 80, min\_len = 10 \\
    \bottomrule
    \end{tabular}%
    }
  \label{tab:hyper-sumamary}%
\end{table}%

\section{Experiment}

\subsection{Experimental Settings}

\subsubsection{Datasets}
We evaluate Fwd-Prompt on the benchmark proposed by \cite{he2023continual}. We consider two MLLMs, InstructBLIP (FlanT5XL) and BLIP2 (FlanT5XL) \cite{li2023blip} for continual learning. When using InstructBLIP and BLIP2 \cite{li2023blip} as MLLMs, the default task order for continual learning is summarized in Table~\ref{tab:task_order_detailed}. Additionally, we expect InstructBLIP to preserve the pre-trained knowledge obtained during joint instruction-tuning before continual learning. Therefore, apart from the new tasks, we also evaluate InstructBLIP on six pre-trained tasks, including Caption COCO \cite{chen2015microsoft}, TextCaps \cite{sidorov2020textcaps}, VQAv2 \cite{goyal2017making}, OKVQA \cite{marino2019ok}, A-OKVQA \cite{schwenk2022okvqa}, and OCR-VQA \cite{mishra2019ocr}. The statistics are provided in Table \ref{tab:dataset_statistics_instructblip} and \ref{tab:dataset_statistics_blip2} and the detailed description of each dataset is in the Appendix.

\subsubsection{Evaluation Metric}
We adopt top-1 accuracy as the metric for VQA-based tasks and CIDEr score for image-captioning tasks. We report three metrics for continual learning, including average accuracy ($\mathcal{A}_t$, higher is better), forgetting (${FGT}_t$, lower is better), and forward transfer (${FWD}_t$, higher is better) after learning task $t$. The definitions of these metrics are provided in the Appendix.

\subsubsection{Hyper-Parameters}
We set the number of key-value pairs $M=20$, the threshold for $K$-rank approximation $\epsilon=0.99$, the threshold for finding matching directions $\theta=0.5$, and the number of selected prompts for input $n_p=5$. Table~\ref{tab:hyper-sumamary} summarizes the hyper-parameters used for training.

\subsubsection{Implementation Details}
We use the implementation of InstructBLIP and BLIP2 provided in the LAVIS library \cite{li2023lavis}. 
We train GQA for 1 epoch and other tasks for 5 epochs with an initial learning rate of 1e-5.
The batch size is set as 32.
In EProj, the projection layer is expanded for each new task, and task keys are learned to select the task ID during inference.
For a fair comparison, Fwd-Prompt follows EProj and additionally learns new projection layers and task keys.
Due to prompt tuning's flexibility, Fwd-Prompt only trains the top 3 layers in Q-Former rather than the entire Q-Former, as it does in InstructBLIP.
We sample 100 instances for each dataset in task $0$ of InsturctBLIP to obtain the pre-trained and residual space.
For BLIP2, we obtain the pre-trained and residual space on the input embeddings of task $1$.
We use the same instruction templates in InstructBLIP \cite{dai2023instructblip} for training.
We use the instruction templates summarized in Table~\ref{tab:prompt-summary} for inference.
We only evaluate BLIP2 on the 5 tasks learned during continual learning since BLIP2 has not been instruction-tuned on multiple datasets.

\subsubsection{Baseline Methods}
We consider 6 baselines for continual learning, including sequential instruction-tuning (SeqIT), EWC \cite{kirkpatrick2017overcoming}, EWC\_TIR \cite{he2023continual}, AGEM \cite{chaudhry2019efficient}, experience replay (ER), and EProj \cite{he2023continual}. EProj is the current state-of-the-art method for MCIT. Additionally, we compare two \emph{non-continual learning} baselines: InstructBLIP before continual learning (Initial) and instruction-tuned on each new task (DirectIT). SeqIT and DirectIT are the lower and upper bounds for continual learning. We store 1\% of old samples for rehearsal-based methods: ER, EProj, and AGEM. The detailed descriptions of the baseline methods are in the Appendix.

\begin{table}[!t]
  \centering
  \caption{Comparison with SOTA methods for MCIT. The average accuracy after learning the final task is reported, i.e., $\mathcal{A}_4$. The backbone model is \emph{InstructBlip}. The average accuracy of each incremental step and forgetting and forward transfer after learning all tasks are reported in Figure~\ref{fig:average_acc_fgt_fwd_instructblip}. The best results are \emph{bold}, and the second best results are \underline{underlined}. ** represents that the P-value is smaller than 0.01.}
  \resizebox{0.99\linewidth}{!}{
    \begin{tabular}{l|cc|ccccccc}
    \toprule
    \multirow{2}[4]{*}{Datasets} & \multicolumn{2}{c|}{Non-CL Baselines} & \multicolumn{7}{c}{CL Baselines} \\
\cmidrule{2-10}          & Initial & DirectIT & SeqIT & EWC   & EWC\_TIR & AGEM  & ER    & EProj & \textbf{Fwd-Prompt} \\
    \midrule
    Caption Coco & 140.81  & 140.81  & 115.85  & 128.27  & 130.77  & 121.89  & \underline{135.00} & 134.99  & \textbf{137.58} \\
    TextCaps & 127.08  & 127.08  & 86.02  & 120.15  & 121.71  & 114.72  & 122.75  & \textbf{126.75} & \underline{125.50} \\
    VQAv2 & 73.47  & 73.47  & 61.08  & 71.54  & \underline{71.96} & 68.31  & 67.06  & 68.03  & \textbf{72.98} \\
    OKVQA & 48.84  & 48.84  & 30.55  & 44.77  & \underline{45.44} & 40.76  & 37.49  & 41.12  & \textbf{48.68} \\
    AOKVQA & 55.31  & 55.31  & 38.52  & 53.16  & \underline{53.51} & 44.88  & 45.68  & 51.85  & \textbf{53.97} \\
    OCR-VQA & 62.37  & 62.37  & 57.84  & 60.28  & \underline{62.30} & 60.37  & 61.38  & \textbf{62.36} & 61.78 \\
    Flickr30k & 84.87  & 97.74  & 79.91  & 92.10  & 94.70  & 90.68  & 95.98  & \textbf{97.99} & \underline{97.53} \\
    VizWiz & 30.65  & 64.58  & 21.54  & 31.14  & 33.53  & 37.83  & \underline{46.36} & 45.05  & \textbf{62.79} \\
    TextVQA & 44.72  & 51.10  & 42.28  & 48.09  & \underline{49.38} & 47.25  & 49.19  & 47.58  & \textbf{50.08} \\
    GQA   & 48.38  & 59.19  & 58.92  & 50.26  & 51.45  & 56.31  & \underline{59.02} & 54.06  & \textbf{60.52} \\
    \midrule
    Average & 71.65 & 78.05 & 59.25  & 69.98  & 71.48  & 68.30  & 71.99  & \underline{72.97} & \textbf{77.14}** \\
    Improvement & /     & /     & 0.00  & +10.73 & +12.22 & +9.05 & +12.74 & \underline{+13.73} & \textbf{+17.89} \\
    \midrule
    Trainable Params & /     & /     & 187M  & 187M  & 187M  & 187M  & 187M  & 189M  & \textbf{48.5M} \\
     Use Old Samples & /     & /     & \XSolidBrush & \XSolidBrush & \XSolidBrush & \Checkmark & \Checkmark & \Checkmark & \XSolidBrush \\
    \bottomrule
    \end{tabular}%
    }
  \label{tab:sota_instructblip}%
\end{table}%

\begin{figure}[!t]
    \centering

    \subfloat[]{
        \includegraphics[width=0.99\linewidth]{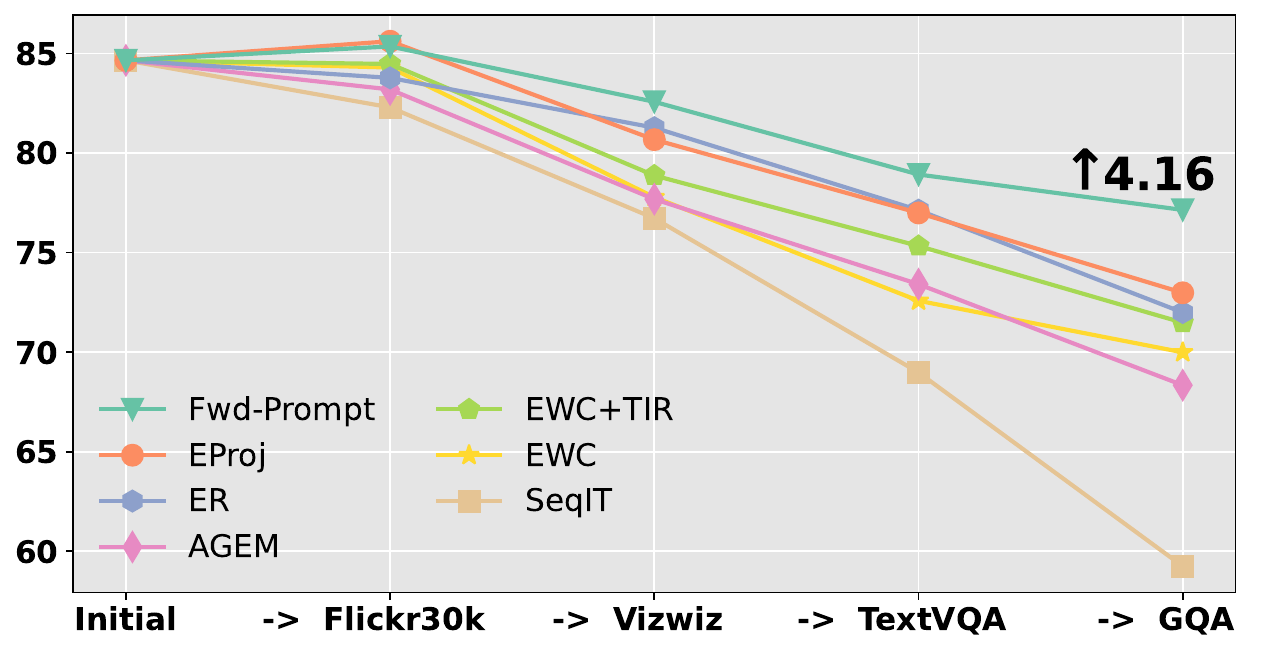}
    }
    
    \subfloat[]{
        \includegraphics[width=0.49\linewidth]{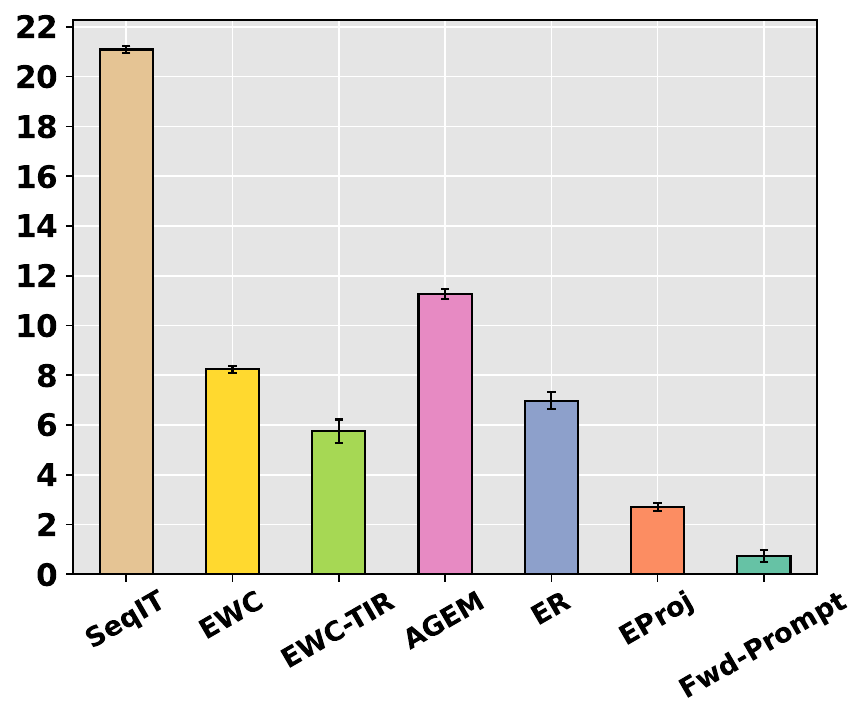}
    }
    \subfloat[]{
        \includegraphics[width=0.49\linewidth]{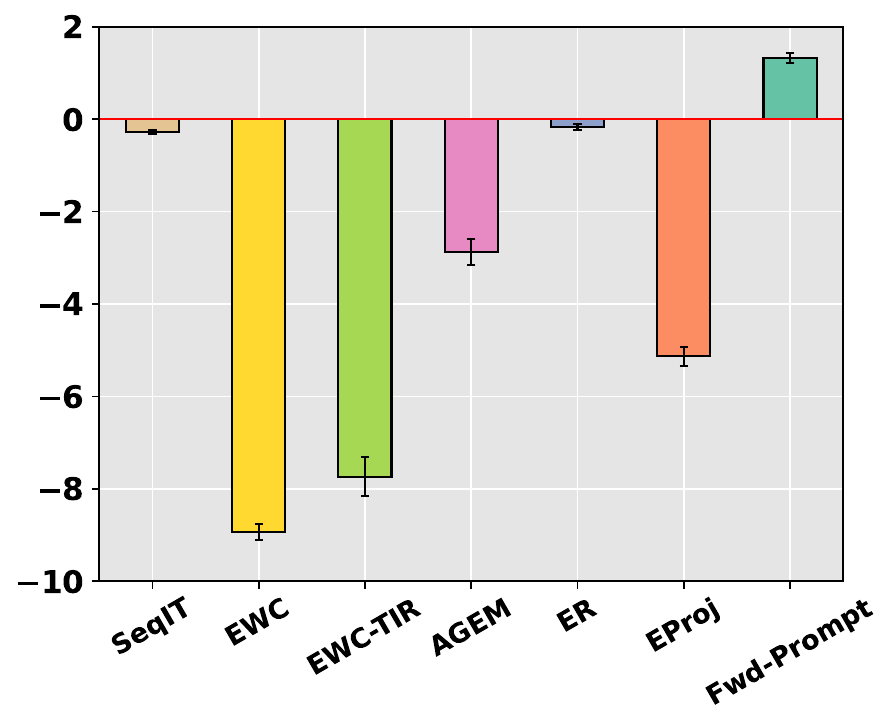}
    }

    \caption{Comparison with SOTA methods in terms of average accuracy, forgetting, and forward transfer. The backbone model is \emph{InstructBLIP}. (a) Average accuracy after learning task $t$, $\mathcal{A}_t (\mathbf{\uparrow})$. (b) Forgetting after learning 4 tasks, ${FGT}_4 (\mathbf{\downarrow})$. (c) Forward transfer after learning 4 tasks, ${FWD}_4 (\mathbf{\uparrow})$. The error bar represents one standard deviation.}
    \label{fig:average_acc_fgt_fwd_instructblip}
\end{figure}

\begin{table}[htbp]
  \centering
  \caption{Comparison with SOTA methods for MCIT. The average accuracy after learning the final task, i.e., $\mathcal{A}_5$, is reported. The backbone model is \emph{BLIP2}. The average accuracy of each incremental step and forgetting and forward transfer after learning all tasks are reported in Figure~\ref{fig:average_acc_fgt_fwd_blip2}. ``ER'' and ``EProj'' require storing 1\% of old samples for rehearsal. The best results are \textbf{bold}, and the second best results are \underline{underlined}. ** represents that the P-value is smaller than 0.01.}
  \resizebox{0.99\linewidth}{!}{
    \begin{tabular}{l|cc|ccccc}
    \toprule
    \multirow{2}[4]{*}{Datasets} & \multicolumn{2}{c|}{Non-CL Baselines} & \multicolumn{5}{c}{CL Baselines} \\
\cmidrule{2-8}          & Initial & DirectIT & SeqIT & EWC\_TIR & ER    & EProj & \textbf{Fwd-Prompt} \\
    \midrule
    Flickr30k & 80.77  & 99.27  & 36.12  & 66.00  & 94.35  & \textbf{98.14 } & \underline{98.09} \\
    TextCaps & 72.04  & 130.67  & 26.96  & 68.35  & 123.85  & \underline{126.69} & \textbf{128.65 } \\
    VQAv2 & 62.79  & 73.44  & 60.75  & 63.61  & 65.97  & \underline{65.98} & \textbf{72.35 } \\
    OCR-VQA & 42.07  & 63.09  & 45.80  & 47.44  & 62.22  & \underline{62.37} & \textbf{62.81 } \\
    GQA   & 43.80  & 56.86  & 58.20  & 49.38  & \underline{58.32} & 50.48  & \textbf{58.80 } \\
    \midrule
    Average & 60.29 & 84.67 & 45.57  & 58.96  & \underline{80.94} & 80.73  & \textbf{84.14}** \\
    Improvement & /     & /     & 0.00  & +13.39 & +35.37 & +35.16 & \textbf{+38.57} \\
    \bottomrule
    \end{tabular}%
    }
  \label{tab:sota_blip2}%
\end{table}%

\begin{figure}[htbp]
    \centering

    \subfloat[]{
        \includegraphics[width=0.9\linewidth]{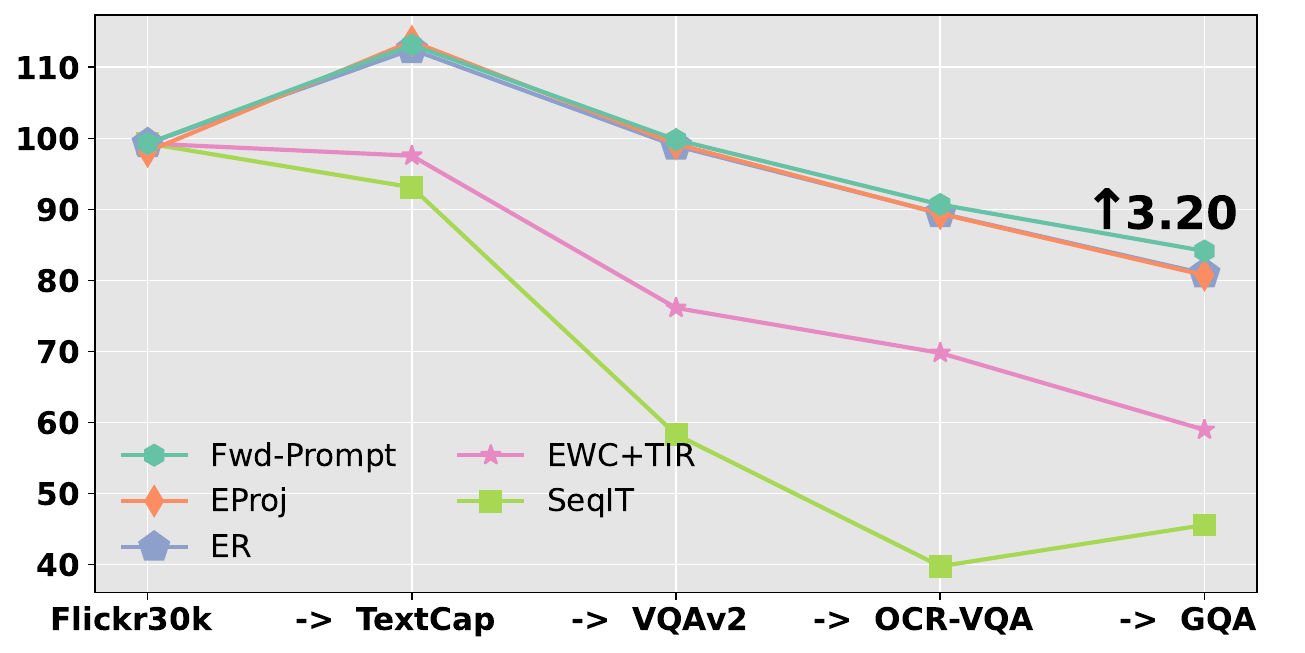}
    }
    
    \subfloat[]{
        \includegraphics[width=0.49\linewidth]{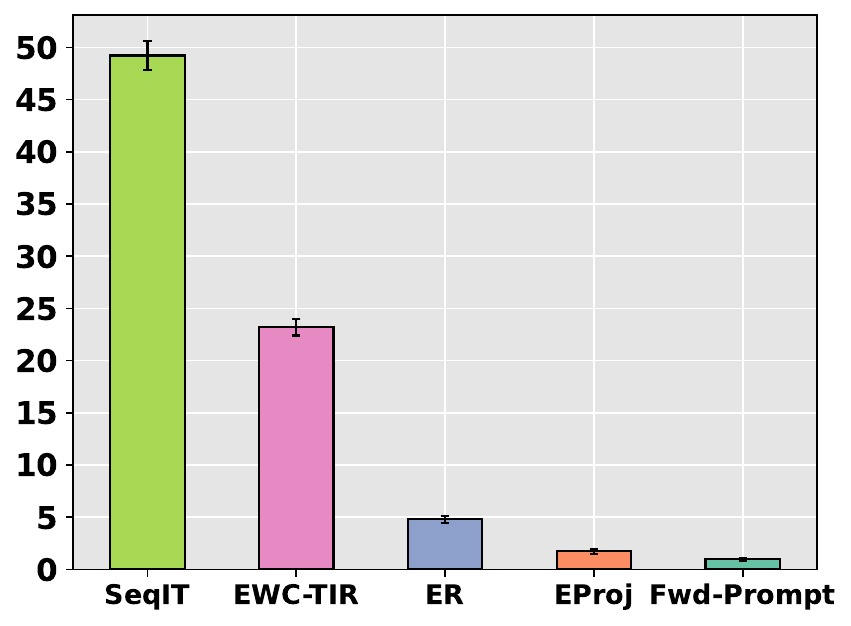}
    }
    \subfloat[]{
        \includegraphics[width=0.49\linewidth]{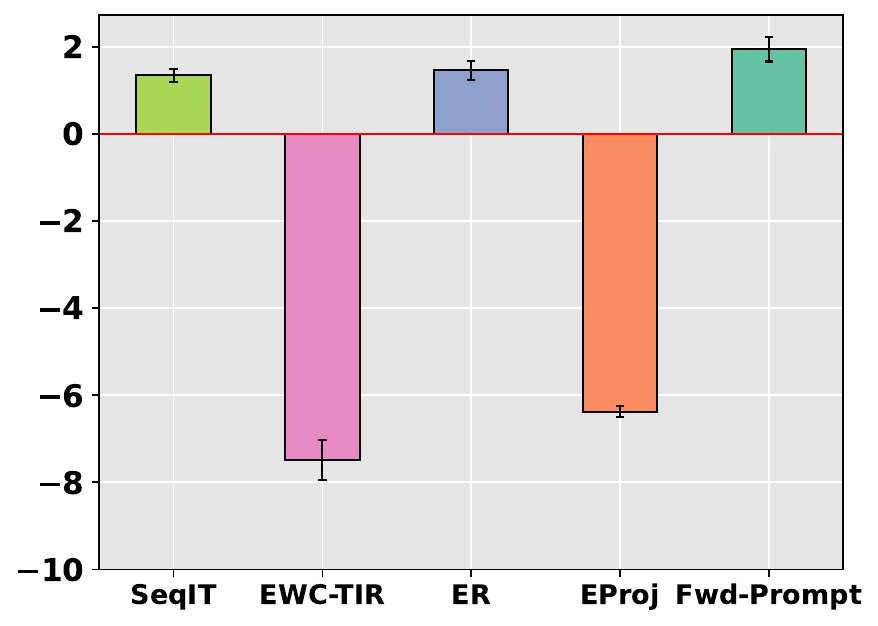}
    }

    \caption{Comparison with SOTA methods in terms of average accuracy, forgetting, and forward transfer. The backbone model is \textbf{BLIP2}. (a) average accuracy after learning task $t$, $\mathcal{A}_t (\mathbf{\uparrow})$. (b) forgetting after learning 5 tasks, ${FGT}_5 (\mathbf{\downarrow})$. (c) forward transfer after learning 5 tasks, ${FWD}_5 (\mathbf{\uparrow})$. The error bar represents one standard derivation.}
    \label{fig:average_acc_fgt_fwd_blip2}
\end{figure}

\subsection{Comparison with SOTA Methods}
Table~\ref{tab:sota_instructblip} and Figure~\ref{fig:average_acc_fgt_fwd_instructblip} highlight the superior performance of Fwd-Prompt when using the InstructBLIP model. Despite not storing old samples and training fewer parameters, Fwd-Prompt surpasses the previous SOTA method by 4.16\%. While EProj is effective at preventing forgetting and thus performs well on pre-trained datasets like OCR-VQA, it does not handle positive forward transfer as effectively as Fwd-Prompt, which excels on both pre-trained and newly added datasets such as GQA.

Regularization-based methods like EWC, EWC\_TIR, and AGEM are significantly impacted by negative forward transfer. Replay-based methods, on the other hand, suffer from either catastrophic forgetting or negative forward transfer. Fwd-Prompt mitigates these issues by leveraging prompt tuning and enhancing positive forward transfer through gradient projection.

The results for BLIP2 are summarized in Table~\ref{tab:sota_blip2} and Figure~\ref{fig:average_acc_fgt_fwd_blip2}. Table~\ref{tab:sota_blip2} shows that EProj performs worse than ER, whereas Fwd-Prompt outperforms ER by 3.20\%. This indicates that Fwd-Prompt is more robust than ER when applied to different MLLMs. Furthermore, Figure~\ref{fig:average_acc_fgt_fwd_blip2} demonstrates that Fwd-Prompt achieves the lowest forgetting and highest forward transfer. Although EProj also reduces catastrophic forgetting, it results in negative forward transfer, decreasing GQA performance from 56.86 to 50.48. In contrast, Fwd-Prompt promotes positive forward transfer, improving GQA performance from 56.86 to 58.80.

\begin{table}[!t]
  \centering
  \caption{The ablation study and hyper-parameter analysis for Fwd-Prompt. The best results are \emph{bold}.}
  \resizebox{0.99\linewidth}{!}{
    \begin{tabular}{clcc}
        \toprule
          & \textbf{Method} & \boldmath{}\textbf{$\mathcal{A}_{4}$}\unboldmath{} & \boldmath{}\textbf{$\Delta$}\unboldmath{} \\
    \midrule
    \multirow{3}[2]{*}{\textbf{Ablation}} & Fwd-Prompt (Ours) & \textbf{77.14} & \textbf{0.00} \\
          & w/o Multimodal Prompt Pool & 63.59 & -13.55 \\
          & w/o Gradient Projection & 74.82 & -2.32 \\
    \midrule
    \multirow{12}[6]{*}{\textbf{Variation}} & Separated Prompt Pools & 74.71 & -2.43 \\
          & More Prompt Tokens ($n_p=10$) & 76.85 & -0.29 \\
          & Larger Prompt Pool ($M=40$) & \textbf{77.31} & \textbf{+0.17} \\
          & Both ($M=40$, $n_p=10$) & 76.89 & -0.25 \\
\cmidrule{2-4}          & Threshold ($\epsilon=0.99$) & \textbf{77.14} & \textbf{0.00} \\
          & Threshold ($\epsilon=0.95$) & 76.50 & -0.64 \\
          & Threshold ($\epsilon=0.90$) & 75.93 & -1.21 \\
          & Threshold ($\epsilon=0.75$) & 75.45 & -1.69 \\
\cmidrule{2-4}          & Threshold ($\theta=0.40$) & 77.11 & -0.03 \\
          & Threshold ($\theta=0.50$) & \textbf{77.14} & \textbf{0.00} \\
          & Threshold ($\theta=0.60$) & 76.97 & -0.17 \\
          & Threshold ($\theta=0.70$) & 76.81 & -0.33 \\
    \bottomrule
    \end{tabular}%
    }
  \label{tab:ablation_hyperparameter_analysis}%
\end{table}

\subsection{Ablation Study and Hyper-Parameter Analysis} 
We consider two ablated versions of Fwd-Prompt. ``w/o Multimodal Prompt Pool'' represents Fwd-Prompt fine-tuning the MLLMs directly without learning a multimodal prompt pool. ``w/o Gradient Projection'' represents Fwd-Prompt updating the prompt without gradient projection. Table~\ref{tab:ablation_hyperparameter_analysis} shows that both the prompt pool and the gradient projection are key to the success of Fwd-Prompt.

We also consider several variations of Fwd-Prompt. ``Separated Prompt Pools'' represents training two prompt pools separately for vision and text input. It amounts to combining the proposed gradient projection with L2P \cite{wang2022learning} naively. The results in Table~\ref{tab:ablation_hyperparameter_analysis} show that ``Separated Prompt Pools'' leads to much worse performance, suggesting that selecting prompts with information from both modalities is crucial. We also try different variations, including learning more prompt tokens, learning a larger prompt pool, and combining both. The results in Table~\ref{tab:ablation_hyperparameter_analysis} show that a larger prompt pool leads to better performance, while prepending more prompt tokens does not. The reason is that more prompt tokens are harder to train.

Additionally, we evaluate Fwd-Prompt with different hyper-parameters $\epsilon$ and $\theta$. The results in Table~\ref{tab:ablation_hyperparameter_analysis} indicate that Fwd-Prompt performs best when $\epsilon=0.99$ and $\theta=0.5$. The results show that Fwd-Prompt is relatively robust when $\theta \geq 0.4$ and $\theta \leq 0.6$. The performance decreases when $\epsilon$ becomes smaller, indicating that precisely estimating the core space is key to the success of Fwd-Prompt. It is reasonable to set $\epsilon=0.99$ since the core space can be effectively approximated with SVD.

\begin{table}[!t]
  \centering
  \caption{The average accuracy after learning the final task under different task orders. The task order is provided in Table~\ref{tab:task_order_detailed}. The best results are \emph{bold}.}
  \resizebox{0.99\linewidth}{!}{
    \begin{tabular}{clcccc}
    \toprule
    \textbf{Backbone} &       & \textbf{SeqIT} & \textbf{ER} & \textbf{EProj} & \textbf{Fwd-Prompt} \\
    \midrule
    \multirow{6}[4]{*}{\textbf{InstructBLIP}} & \textbf{Order 1} & 59.25  & 71.99  & 72.97  & \textbf{77.14} \\
          & \textbf{Order 2} & 57.84  & 71.60  & 72.51  & \textbf{76.89} \\
          & \textbf{Order 3} & 59.13  & 72.85  & 73.06  & \textbf{77.54} \\
          & \textbf{Order 4} & 46.27  & 68.85  & 69.47  & \textbf{75.36} \\
\cmidrule{2-6}          & \textbf{Average} & 55.62  & 71.32  & 72.00  & \textbf{76.73} \\
          & \boldmath{}\textbf{$\Delta$}\unboldmath{} & /     & +15.70 & +16.38 & \textbf{+21.11} \\
    \midrule
    \multirow{5}[4]{*}{\textbf{BLIP2}} & \textbf{Order 5} & 45.57  & 80.94  & 80.73  & \textbf{84.14} \\
          & \textbf{Order 6} & 44.86  & 77.53  & 78.12  & \textbf{82.56} \\
          & \textbf{Order 7} & 44.53  & 77.46  & 76.41  & \textbf{81.62} \\
\cmidrule{2-6}          & \textbf{Average} & 44.99  & 78.64  & 78.42  & \textbf{82.77} \\
          & \boldmath{}\textbf{$\Delta$}\unboldmath{} & /     & +33.65 & +33.43 & \textbf{+37.78} \\
    \bottomrule
    \end{tabular}%
    }
  \label{tab:exp_diff_task_order}%
\end{table}

\begin{table}[!t]
  \centering
  \caption{The task orders used in the experiments. * represents the default task order for each backbone MLLM.}
  \resizebox{0.99\linewidth}{!}{
    \begin{tabular}{lll}
    \toprule
    \textbf{Backbone} & \textbf{Task Order} & \textbf{Task Sequence} \\
    \midrule
    \multirow{4}[8]{*}{\textbf{InstructBLIP}} & \textbf{Order 1*} & Flickr30K $\rightarrow$ VizWiz $\rightarrow$ TextVQA $\rightarrow$ GQA \\
\cmidrule{2-3}          & \textbf{Order 2} & GQA $\rightarrow$ TextVQA $\rightarrow$ VizWiz $\rightarrow$ Flickr30K \\
\cmidrule{2-3}          & \textbf{Order 3} & TextVQA $\rightarrow$ GQA $\rightarrow$ Flickr30K $\rightarrow$ VizWiz \\
\cmidrule{2-3}          & \textbf{Order 4} & \makecell{Flickr30K(S1) $\rightarrow$ VizWiz(S1) $\rightarrow$ TextVQA(S1) $\rightarrow$ GQA(S1) $\rightarrow$\\
                          Flickr30K(S2) $\rightarrow$ VizWiz(S2) $\rightarrow$ TextVQA(S2) $\rightarrow$ GQA(S2) $\rightarrow$\\
                          Flickr30K(S3) $\rightarrow$ VizWiz(S3) $\rightarrow$ TextVQA(S3) $\rightarrow$ GQA(S3)} \\
    \midrule
    \multirow{3}[6]{*}{\textbf{BLIP2}} & \textbf{Order 5*} & Flickr30K $\rightarrow$ TextCaps $\rightarrow$ VQAv2 $\rightarrow$ OCR-VQA $\rightarrow$ GQA \\
\cmidrule{2-3}          & \textbf{Order 6} & GQA $\rightarrow$ OCR-VQA $\rightarrow$ VQAv2 $\rightarrow$ TextCaps $\rightarrow$ Flickr30K \\
\cmidrule{2-3}          & \textbf{Order 7} & OCR-VQA $\rightarrow$ GQA $\rightarrow$ VQAv2 $\rightarrow$ Flickr30K $\rightarrow$ TextCaps \\
    \bottomrule
    \end{tabular}%
    }
  \label{tab:task_order_detailed}%
\end{table}

\subsection{Experiments on Different Task Orders}

We consider 6 different task orders for continual learning to further evaluate Fwd-Prompt. The task sequence for each task order is summarized in Table~\ref{tab:task_order_detailed}. The results in Table~\ref{tab:exp_diff_task_order} show that Fwd-Prompt consistently outperforms the previous SOTA method, EProj. Specifically, Fwd-Prompt achieves an average performance of 76.73\% and 82.77\% on InstructBLIP and BLIP2 across different task orders, while EProj achieves 72.00\% and 78.42\%, respectively. The results also indicate that Fwd-Prompt is more robust to task orders than ER and EProj.

\begin{table}[!t]
  \centering
  \caption{The average accuracy after learning the final task when models are trained on more tasks. The task order is Order 4, defined in Table \ref{tab:task_order_detailed}. The best results are \emph{bold}.}
  \resizebox{0.99\linewidth}{!}{
    \begin{tabular}{ccccc}
    \toprule
          & \textbf{SeqIT} & \textbf{ER} & \textbf{EProj} & \textbf{Fwd-Prompt} \\
    \midrule
    \textbf{Caption Coco} & 75.75  & 114.53  & 107.12  & \textbf{124.24} \\
    \textbf{TextCaps} & 66.35  & 116.83  & 118.39  & \textbf{125.29} \\
    \textbf{VQAv2} & 47.14  & 65.66  & 66.05  & \textbf{72.92} \\
    \textbf{OKVQA} & 24.34  & 35.52  & 34.76  & \textbf{47.45} \\
    \textbf{AOKVQA} & 23.76  & 45.75  & 48.82  & \textbf{52.30} \\
    \textbf{OCR-VQA} & 41.60  & 58.48  & 59.34  & \textbf{61.04} \\
    \textbf{Flickr30k} & 61.11  & 96.04  & 97.05  & \textbf{97.44} \\
    \textbf{VizWiz} & 21.54  & 47.16  & 53.14  & \textbf{61.68} \\
    \textbf{TextVQA} & 42.20  & 48.83  & 50.03  & \textbf{50.57} \\
    \textbf{GQA} & 58.92  & 59.68  & 59.98  & \textbf{60.68} \\
    \midrule
    \textbf{Average} & 46.27  & 68.85  & 69.47  & \textbf{75.36} \\
    \bottomrule
    \end{tabular}%
    }
  \label{tab:exp_more_task}%
\end{table}

\subsection{Experiments on More Tasks}

To evaluate Fwd-Prompt on more tasks, we randomly divide the training set of Flickr30k, VizWiz, TextVQA, and GQA into three splits: S1, S2, and S3, respectively. The task order is denoted as ``Order 4'' and shown in Table~\ref{tab:task_order_detailed}. We note that the dev sets of Flickr30k, VizWiz, TextVQA, and GQA remain unchanged. In this setting, the forgetting in pre-trained tasks is exacerbated and thus more challenging. The results in Table~\ref{tab:exp_more_task} indicate that Fwd-Prompt outperforms existing methods by a large margin. Existing methods suffer from catastrophic forgetting on pre-trained tasks such as VQAv2, while Fwd-Prompt maintains the most pre-trained performance.

\begin{figure}[!t]
    \centering
    \subfloat[]{
        \includegraphics[width=0.49\linewidth]{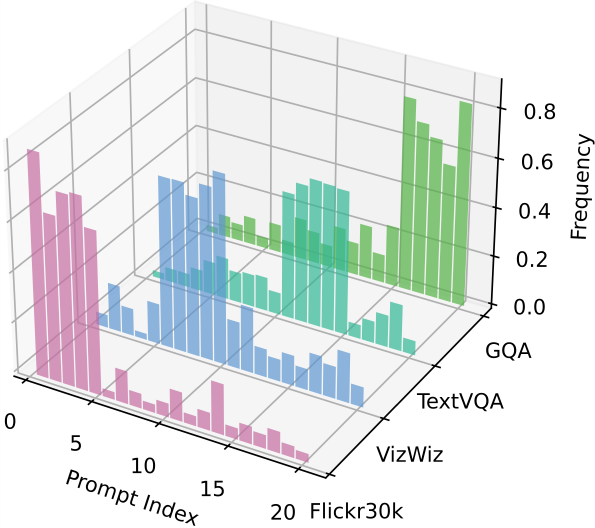}
        \label{fig:forward_transfer_hist_a}
    }
    \subfloat[]{
        \includegraphics[width=0.49\linewidth]{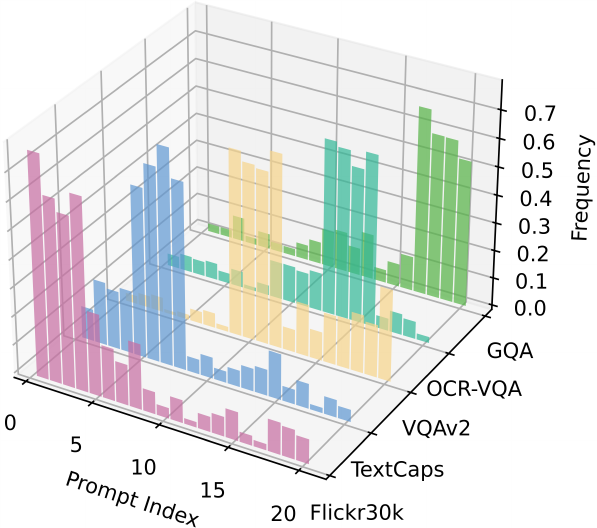}
        \label{fig:forward_transfer_hist_b}
    }
    \caption{The histogram of prompt selection frequency in Fwd-Prompt. (a) The backbone model is InstructBLIP. (b) The backbone model is BLIP2.}
    \label{fig:forward_transfer_hist}
\end{figure}

\begin{figure}[!t]
    \centering
        \includegraphics[width=0.70\linewidth]{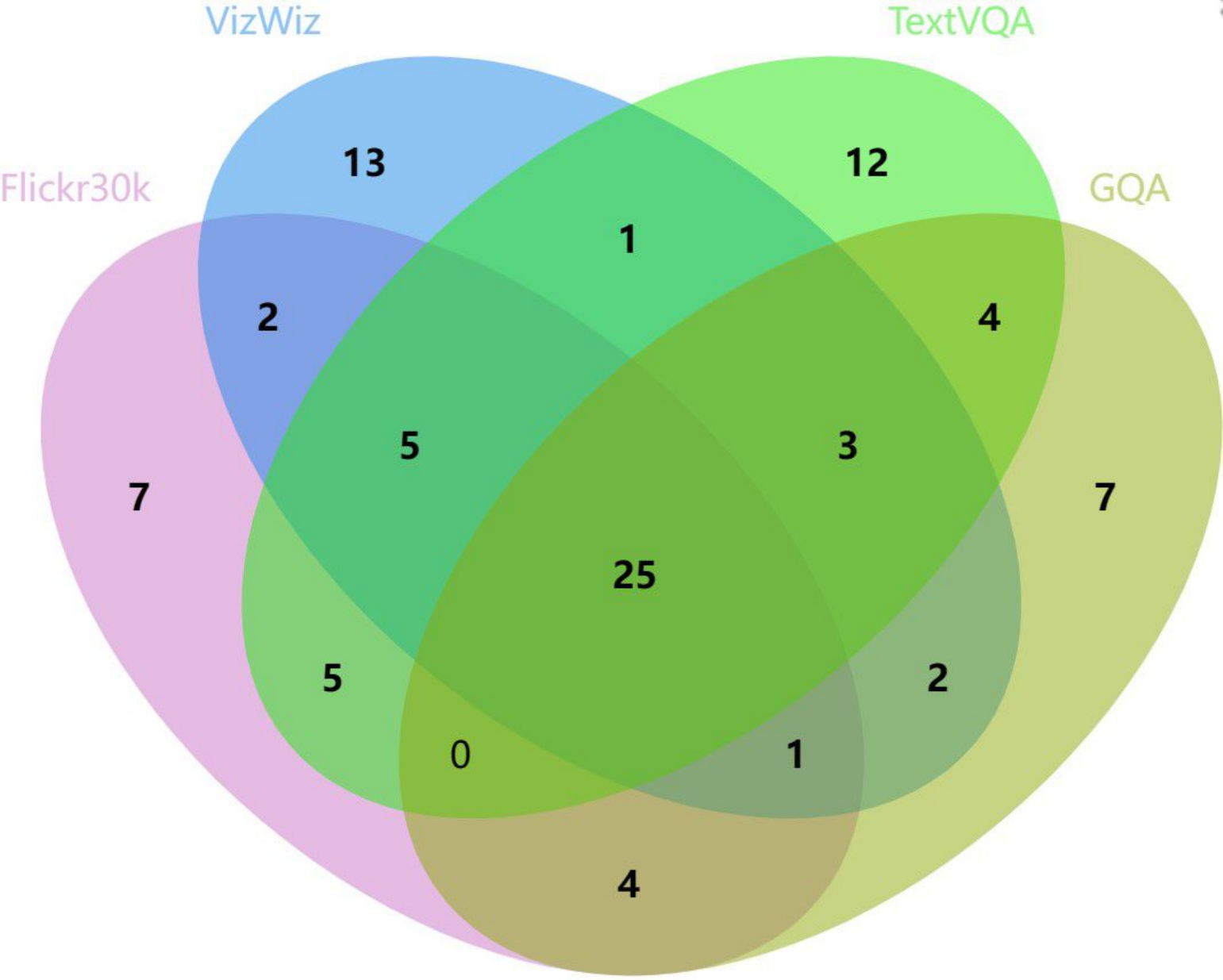}
        \label{fig:forward_transfer_venn_diagram_a}
        \hspace{0.5cm}
        \includegraphics[width=0.05\linewidth]{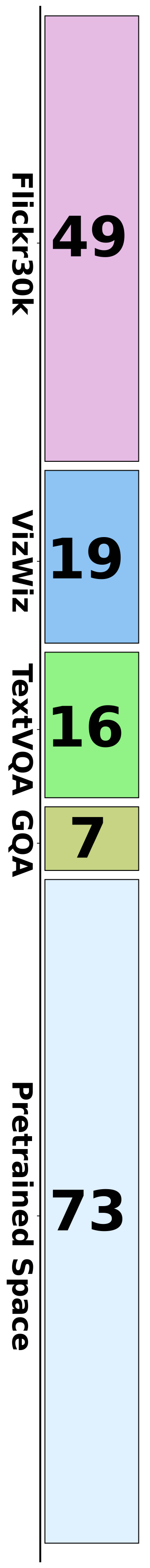}
        \label{fig:forward_transfer_venn_diagram_b}
    \caption{The subspace allocation in Fwd-Prompt. The numbers represent the number of column vectors (i.e., directions). The backbone model is InstructBLIP. Left: The Venn diagram shows the conflicting space of each task. Right: The number of \emph{unique} column vectors in the conflicting space for each task.}
    \label{fig:forward_transfer_venn_diagram}
\end{figure}

\subsection{Visualization of Forward Transfer in Fwd-Prompt}

We visualize the prompt selection frequency for each dataset in Figure~\ref{fig:forward_transfer_hist}. The visualization shows that each task has a higher frequency of selecting a non-overlapping set of prompts. It indicates that the model learns task-specific information and learns to identify the task during training. Our finding is consistent with L2P \cite{wang2022learning}. Additionally, we observe that interaction across tasks exists. For example, in Figure \ref{fig:forward_transfer_hist_b}, when the task is an image captioning task like Flickr30k, the model also selects prompts from index 1-4, which corresponds to another image captioning task, TextCaps. In this case, the knowledge transfer is enhanced by selecting prompts from different tasks.

We also visualize the subspace allocation in Fwd-Prompt in Figure~\ref{fig:forward_transfer_venn_diagram}. Figure~\ref{fig:forward_transfer_venn_diagram_a} shows considerable overlap in the conflicting space of different tasks. Fwd-Prompt avoids conflict between tasks by allocating different conflicting spaces to different tasks. Figure~\ref{fig:forward_transfer_venn_diagram_b} visualizes the number of \emph{unique} column vectors in the conflicting space for each task. For example, when the model learns the first task, Flickr30k, the unique directions in conflicting space 1 are 49 (=7+5+0+4+1+25+5+2). When the model learns the second task, VizWiz, the unique directions in conflicting space 2 are 19 (=13+1+3+2). By identifying conflicting spaces, Fwd-Prompt avoids interference between new tasks and thus achieves anti-forgetting. Furthermore, Fwd-Prompt enhances positive forward transfer by updating prompts in the pre-trained space, which contains the core space of unseen tasks.

\begin{table}[!t]
  \centering
  \caption{Comparison with SOTA methods in training time of continual learning. The backbone model is \emph{InstructBLIP}.}
  \resizebox{0.60\linewidth}{!}{
    \begin{tabular}{lcc}
    \toprule
    \multicolumn{1}{c}{\textbf{Method}} & \textbf{Run Time (min)} & \textbf{Ratio} \\
    \midrule
    SeqIT & 502   & $\times 1.0$ \\
    EWC   & 745   & $\times 1.5$ \\
    A-GEM & 1047  & $\times 2.1$ \\
    EProj & 972   & $\times 1.9$ \\
    EWC\_TIR & 768   & $\times 1.5$ \\
    ER    & 945   & $\times 1.9$ \\
    \midrule
    Fwd-Prompt & 660   & $\times 1.3$ \\
    \bottomrule
    \end{tabular}%
    }
  \label{tab:runtime_comparison}%
\end{table}

\subsection{Run Time Analysis} \quad
The comparison of training time is shown in Table~\ref{tab:runtime_comparison}. The baseline method, SeqIT, sets a benchmark with a runtime of 502 minutes. Methods like EWC and EWC\_TIR exhibit a 50\% increase in runtime, indicating a moderate efficiency cost for potentially enhanced learning stability or performance. On the higher end, A-GEM and EProj almost double the training time of SeqIT, which could be attributed to their more complex mechanisms for managing catastrophic forgetting. Notably, Fwd-Prompt shows a promising balance, only running 30\% longer than SeqIT, demonstrating its superior practicability in real-world scenarios.

\begin{table*}[htbp]
  \centering
  \caption{The qualitative analysis of MCIT. We randomly selected four examples from Caption Coco, Flickr30k, Vizwiz, and OCR-VQA (left to right) after MCIT. We compare the responses of ER, EProj, and Fwd-Prompt.}
  \resizebox{0.99\linewidth}{!}{
    \begin{tabular}{p{2cm}|p{3cm}|p{3cm}|p{3cm}|p{3cm}}
    \toprule
    \textbf{Input Image} & \begin{minipage}[b]{0.30\columnwidth}
		\centering
		\raisebox{-.5\height}{\includegraphics[width=0.6\linewidth]{image/case_study/1.pdf}}
	\end{minipage}
     &  \begin{minipage}[b]{0.35\columnwidth}
		\centering
		\raisebox{-.5\height}{\includegraphics[width=\linewidth]{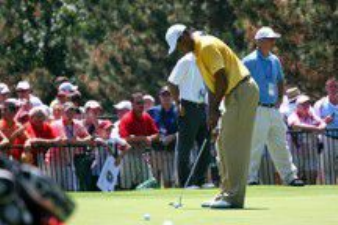}}
	\end{minipage} & \begin{minipage}[b]{0.33\columnwidth}
		\centering
		\raisebox{-.5\height}{\includegraphics[width=0.6\linewidth]{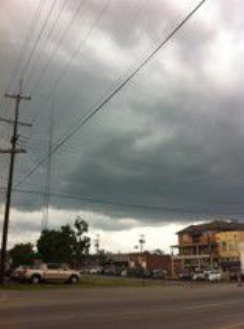}}
	\end{minipage}  & \begin{minipage}[b]{0.35\columnwidth}
		\centering
		\raisebox{-.5\height}{\includegraphics[width=\linewidth]{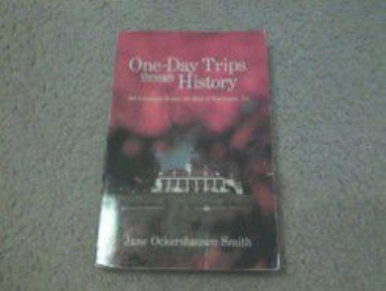}}
	\end{minipage} \\
    \midrule
    \textbf{Input Instruction} & A short image description: & A short image description: & Question: I was wondering how ominous the sky is looking here at the northwest. Short answer: & What is the title of this book? \\
    \midrule
    \textbf{ER Response} & A police bus travels in its designated lane on a highway. & A man focuses on his putt as spectators watch eagerly in the background. & cloudy & One-Day Trips through History \\
    \midrule
    \textbf{EProj Response} & A white bus is driving on a highway leading to a turn & A man wearing a yellow shirt and light brown slacks plays golf.  & unanswerable & One-Day Trips through History: 200 Excursions Within 150 Miles of Washington, D.C. \\
    \midrule
    \textbf{Fwd-Prompt Response} & A white and red bus drives on a busy four-lane road with other vehicles & A golfer in a yellow shirt and white cap is preparing to putt on the green at a golf course, with an attentive audience and a caddie in the background. & dark cloudy & One-Day Trips through History: 200 Excursions Within 150 Miles of Washington, D.C. \\
    \midrule
    \textbf{Ground Truth} & A bus decorated in white and red was traveling along the busy four-lane road with four cars. & A guy in a yellow shirt and light brown slacks is playing golf while the crowd watches. & cloudy & One-Day Trips through History: 200 Excursions Within 150 Miles of Washington, D.C. \\
    \bottomrule
    \end{tabular}%
    }
  \label{tab:case_study}
\end{table*}

\subsection{Qualitative Analysis} \quad 

In Table~\ref{tab:case_study}, we compare three MCIT methods—ER, EProj, and our proposed Fwd-Prompt—using the InstructBlip model across four different images from Caption Coco, Flickr30k, Vizwiz, and OCR-VQA. Fwd-Prompt consistently delivers more detailed and context-rich responses, proving its effectiveness in handling diverse multimodal tasks. For example, in the first image, ER mentions a ``police bus'', EProj describes the bus's color and direction, and Fwd-Prompt provides a detailed scene that closely matches the ground truth. In the golf scene, Fwd-Prompt outshines the others by detailing not just the golfer's actions but also the setting and audience, aligning closely with the actual scene. This indicates that Fwd-Prompt better preserves knowledge gained before and after continual learning.

\section{Conclusion}
In this paper, we analyze the input embeddings and reveal that the conflict in input embeddings leads to forgetting and negative forward transfer. To address this, we present Fwd-Prompt, a novel approach to addressing the critical challenges of catastrophic forgetting and negative forward transfer in MCIT. The experimental results show that Fwd-Prompt outperforms SOTA methods significantly while requiring less training time, fewer trainable parameters, and no old samples. This research sheds light on the potential continual learning ability of MLLMs and encourages future studies to explore MCIT.

\bibliography{reference}
\bibliographystyle{IEEEtran}

\begin{IEEEbiography}[{\includegraphics[width=1in,height=1.25in,clip,keepaspectratio]{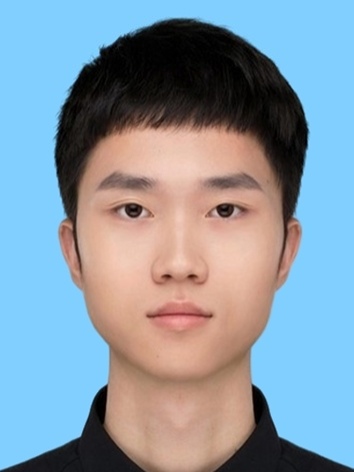}}]{Junhao Zheng}
received his undergraduate degree in computer science from the South China University of Technology, Guangzhou, China, in 2022. He is currently pursuing his Ph.D. degree at the School of Computer Science and Engineering, South China University of Technology, Guangzhou, China. His current research interests include continual learning and large language models.
\end{IEEEbiography}

\begin{IEEEbiography}[{\includegraphics[width=1in,height=1.25in,clip,keepaspectratio]{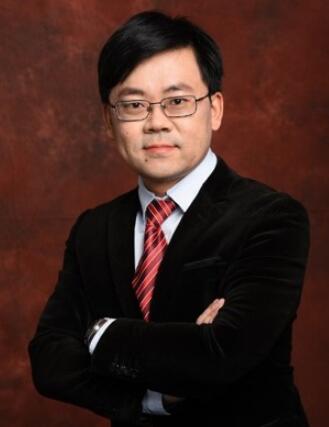}}]{Qianli Ma}
(Member, IEEE) received his Ph.D. degree in computer science from the South China University of Technology, Guangzhou, China, in 2008. He is a Professor at the School of Computer Science and Engineering, South China University of Technology. From 2016 to 2017, he was a Visiting Scholar at the University of California, San Diego. His current research interests include machine learning algorithms, data-mining methodologies, and their applications.
\end{IEEEbiography}

\begin{IEEEbiography}[{\includegraphics[width=1in,height=1.25in,clip,keepaspectratio]{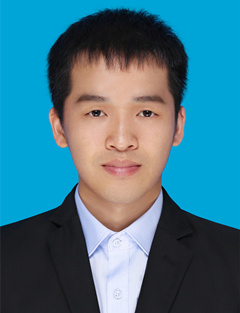}}]{Zhen Liu}
received his bachelor’s degree in software engineering from South-Central Minzu University, Wuhan, China, in 2018. He is currently pursuing his Ph.D. degree at the School of Computer Science and Engineering, South China University of Technology, Guangzhou, China. His research interests include machine learning, deep learning, and time-series analysis.
\end{IEEEbiography}

\begin{IEEEbiography}[{\includegraphics[width=1in,height=1.25in,clip,keepaspectratio]{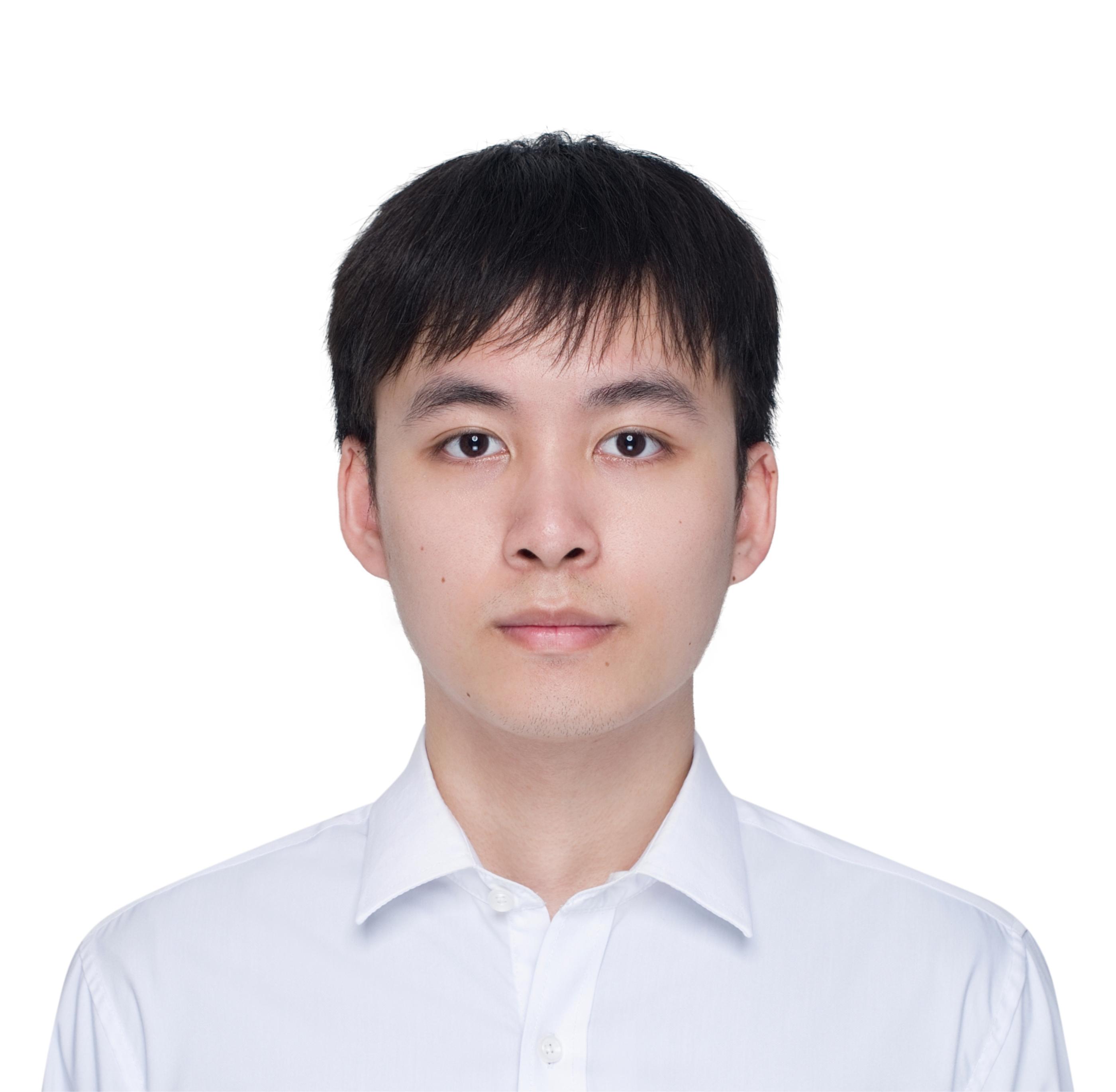}}]{Binquan Wu}
is currently a Ph.D. candidate at the School of Computer Science and Engineering, South China University of Technology. He obtained his B.E. degree from South China University of Technology in 2019. His research interests include data mining, user behavior modeling, and self-supervised learning. He has published a paper at the prestigious conference ICDE.
\end{IEEEbiography}

\begin{IEEEbiography}[{\includegraphics[width=1in,height=1.25in,clip,keepaspectratio]{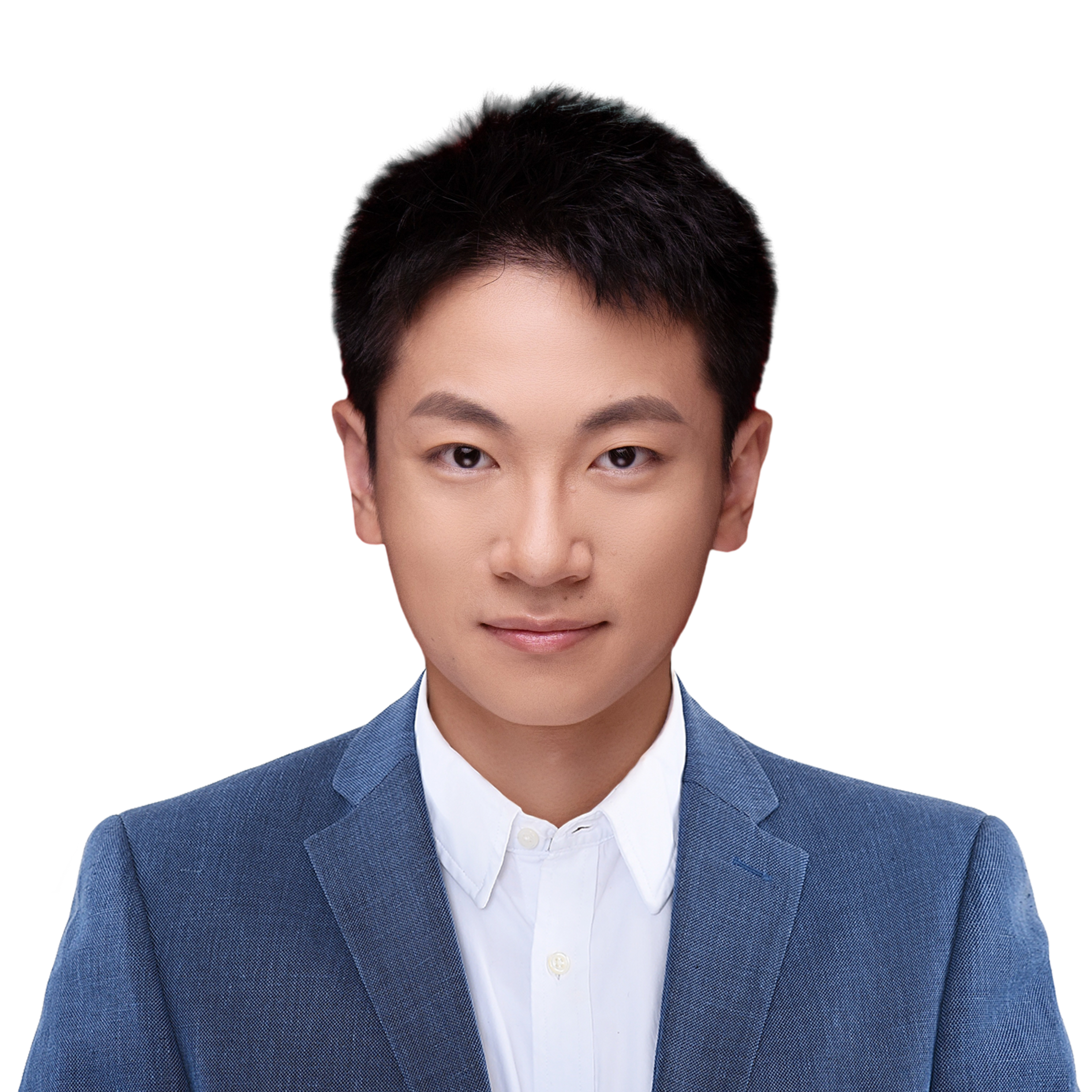}}]{Huawen Feng}
received his undergraduate degree in computer science from the South China University of Technology, Guangzhou, China, in 2021. He is currently pursuing his doctorate and is interested in knowledge alignment and large language models.
\end{IEEEbiography}


\appendix  

\section*{Datasets}
The detailed description of the datasets in this paper is as follows:
\begin{itemize}
    \item Caption COCO \cite{chen2015microsoft}: The Microsoft Common Objects in Context (MS COCO) dataset is used extensively for object detection, segmentation, key-point detection, and image captioning. This dataset includes 328,000 images. We adopt the Karpathy split \cite{karpathy2015deep}, which partitions the dataset into 82,000 training images, 5,000 validation images, and 5,000 test images.
    \item TextCaps \cite{sidorov2020textcaps}: TextCaps demands that models interpret and contemplate the text within images to create descriptive captions. These models must integrate the unique aspect of textual content within the images, combining it with the visual elements to produce descriptions of the images. 
    \item VQAv2 \cite{goyal2017making}: VQAv2 is a dataset containing open-ended questions about images. These questions require an understanding of vision, language, and commonsense knowledge.
    \item OKVQA \cite{marino2019ok}: OK-VQA is designed for visual question answering, demanding techniques that utilize external knowledge sources, such as Wikipedia, for responding. It includes 14,055 open-ended questions, each accompanied by five verifiable answers. The dataset has been meticulously curated to guarantee that every question necessitates external knowledge for accurate answers. Additionally, questions with frequently recurring answers have been limited to minimize bias in the dataset.
    \item A-OKVQA \cite{schwenk2022okvqa}: A-OKVQA is a crowdsourcing dataset consisting of approximately 25,000 diverse questions that demand a wide range of commonsense and world knowledge for answers. Unlike other knowledge-based Visual Question Answering (VQA) datasets, the questions in A-OKVQA typically cannot be resolved by merely consulting a knowledge base. Instead, they necessitate a form of commonsense reasoning about the scene shown in the image.
    \item OCR-VQA \cite{mishra2019ocr}: OCR-VQA comprises visual questions demanding that models interpret text within images. It includes 800k training, 100k validation, and 100k testing instances.
    \item Flickr30k \cite{young2014image}: The Flickr30k dataset comprises 31,000 images sourced from Flickr, each accompanied by five verified captions. The Flickr30k dataset is a widely used benchmark for evaluating sentence-based image descriptions.
    \item VizWiz \cite{gurari2018vizwiz}: VizWiz is a dataset designed explicitly for the needs of the visually impaired. It comprises images captured by visually impaired individuals, each paired with a spoken question and 10 crowdsourced responses. The dataset challenges AI to accurately interpret and respond to visual questions and recognize when a question is unanswerable.
    \item TextVQA \cite{singh2019towards}: TextVQA demands that models understand the visual text to respond to queries.
    \item GQA \cite{hudson2019gqa}: The GQA dataset is an extensive visual question-answering collection featuring authentic images from the Visual Genome dataset and well-balanced question-answer combinations. Every training and validation image includes detailed scene graph annotations that identify the classes and attributes of objects within the scene and their interrelations. 
\end{itemize}

\section*{Evaluation Metric}
We adopt top-1 accuracy as the metric for VQA-based tasks and CIDEr score for image-captioning tasks.
The CIDEr (Consensus-based Image Description Evaluation) score is a metric used to evaluate the quality of captions generated by image captioning models compared to human-written captions.
We report three metrics for continual learning, including average accuracy ($\mathcal{A}_t$, higher is better), forgetting (${FGT}_t$, lower is better), and forward transfer (${FWD}_t$, higher is better) after learning task $t$.
Specifically, 
\begin{equation}
 \mathcal{A}_t = \frac{1}{t}\Sigma_{i=1}^{t}a_{t,i},
\end{equation}
where $a_{t,i}$ is the performance score of task $i$ after learning task $t$.
The forgetting is computed as the average performance degradation of all learned tasks:
\begin{equation}
    {FGT}_t = \frac{1}{t-1}\Sigma_{i=1}^{t-1}[ max_{j<t}(\{a_{j,i}\}_j) - a_{t,i}]
\end{equation}
The forward transfer is computed as the difference between continual learning and direct instruction-tuning on task $t$: 
\begin{equation}
    {FWD}_t = a_{t,t} - a_{t}
\end{equation}
When ${FWD}_t > 0$, it implies positive forward transfer for learning task $t$. 
When ${FWD}_t < 0$, it implies negative forward transfer for learning task $t$.

\section*{Baselines Methods}

The detailed description of the baseline methods in the experiment is as follows:
\begin{itemize}
    \item Sequential Instruction-Tuning (SeqIT): SeqIT refers to instruction-tuning MLLMs on new tasks sequentially without any constraint to forgetting. SeqIT is the lower bound for continual learning. 
    \item Direct Instruction-Tuning (DirectIT): DirectIT refers to instruction-tuning MLLMs on each task. Unlike the methods for continual learning, DirectIT adapts MLLMs to only \textbf{one} new task. In other words, DirectIT requires training $N$ MLLMs for $N$ new tasks, while continual learning only trains one MLLM for $N$ new tasks. Therefore, DirectIT is the upper bound for continual learning.
    \item Experience Replay (ER): ER is a technique widely used in continual learning to help models retain and integrate old knowledge while learning new tasks. ER stores a subset of old data and jointly trains the model on new and old data when adapting to new tasks. In our experiment, we randomly store 1\% of old instances.
    \item EWC \cite{kirkpatrick2017overcoming}: Elastic Weight Consolidation (EWC) is a technique in continual learning used to prevent catastrophic forgetting in neural networks. It works by selectively slowing down learning on certain neural network weights based on their importance to previously learned tasks. By adding a regularization term to the loss function, EWC penalizes changes to crucial weights, allowing the network to retain old knowledge while continuing to learn new information. In our experiment, the weight of the regularization term is set as 5000.
    \item EWC\_TIR \cite{he2023continual}: EWC\_TIR uses task similarity scores to adaptively weight the regularization term in the learning process. It involves storing task embeddings for each known task and corresponding task IDs for parameters. These apply adaptive importance weighting to the regularization term during training based on the similarity between the current and previous tasks. This method does not increase the inference cost and has a manageable increase in training cost compared to classic regularization-based methods.
    \item AGEM \cite{chaudhry2019efficient}: Averaged Gradient Episodic Memory (A-GEM) in continual learning is an efficient variant of the Gradient Episodic Memory (GEM) model \cite{lopez2017gradient}. A-GEM addresses the computational and memory intensity of GEM by ensuring that the average episodic memory loss over previous tasks does not increase during training. It modifies the training process based on a single constraint derived from a random subset of episodic memory.
    \item EProj \cite{he2023continual}: EProj is the current SOTA method for MCIT. EProj employs task similarity scores to decide whether to add a task-specific module for a new task. The training process involves training both the task-specific module and a task-specific key. During testing, the similarity between the test sample embedding and the task-specific keys is computed to retrieve the task ID with the highest similarity score. 
\end{itemize}

%





\ifCLASSOPTIONcaptionsoff
  \newpage
\fi

\end{document}